\documentclass[final]{cvpr}

\usepackage[pagebackref=true,breaklinks=true,colorlinks,bookmarks=false]{hyperref}

\usepackage{times}
\usepackage{epsfig}
\usepackage{graphicx}
\usepackage{amsmath}
\usepackage{amssymb}
\usepackage{multirow}
\usepackage{subfigure}
\usepackage{bm}
\usepackage{enumitem}
\usepackage{arydshln}
\usepackage{adjustbox}
\usepackage[accsupp]{axessibility}

\graphicspath{{Images/}}

\def\cvprPaperID{3384} 
\def\confYear{CVPR 2022}

\author{ 
Xingyu Chen$^1$\thanks{Corresponding author, chenxingyu@kuaishou.com}\hspace{0.15in} Yufeng Liu$^{3}$\hspace{0.15in} Yajiao Dong$^1$ \hspace{0.15in} Xiong Zhang$^2$\hspace{0.15in} Chongyang Ma$^1$\hspace{0.15in} \\ Yanmin Xiong$^1$\hspace{0.3in} Yuan Zhang$^1$\hspace{0.3in}  Xiaoyan Guo$^1$ \\
$^1$Y-tech, Kuaishou Technology   \hspace{0.3in}   $^2$ YY Live, Baidu Inc. \\  $^3$SEU-ALLEN Joint Center, Institute for Brain and Intelligence, Southeast University, China. \\ 
}

\newcommand{\nothing}[1]{}

\newcommand{\loss}{\mathcal{L}}
\newcommand{\lossmesh}{\loss_{mesh}}

\newcommand{\lossposetwo}{\loss_{pose2D}}

\newcommand{\lossnormal}{\loss_{norm}}
\newcommand{\lossedge}{\loss_{edge}}
\newcommand{\losstotal}{\loss_{total}}
\newcommand{\losscontwo}{\loss_{con2D}}
\newcommand{\lossconthree}{\loss_{con3D}}

\begin{document}

\title{MobRecon: Mobile-Friendly Hand Mesh Reconstruction from Monocular Image}

\maketitle
\pagestyle{empty}  
\thispagestyle{empty} 

\begin{abstract}
    In this work, we propose a framework for single-view hand mesh reconstruction, which can simultaneously achieve high reconstruction accuracy, fast inference speed, and temporal coherence. Specifically, for 2D encoding, we propose lightweight yet effective stacked structures. Regarding 3D decoding, we provide an efficient graph operator, namely depth-separable spiral convolution. Moreover, we present a novel feature lifting module for bridging the gap between 2D and 3D representations. This module begins with a map-based position regression (MapReg) block to integrate the merits of both heatmap encoding and position regression paradigms for improved 2D accuracy and temporal coherence. Furthermore, MapReg is followed by pose pooling and pose-to-vertex lifting approaches, which transform 2D pose encodings to semantic features of 3D vertices. Overall, our hand reconstruction framework, called MobRecon, comprises affordable computational costs and miniature model size, which reaches a high inference speed of 83FPS on Apple A14 CPU. Extensive experiments on popular datasets such as FreiHAND, RHD, and HO3Dv2 demonstrate that our MobRecon achieves superior performance on reconstruction accuracy and temporal coherence. 
    Our code is publicly available at \url{https://github.com/SeanChenxy/HandMesh}.
\end{abstract}

\section{Introduction}

Single-view hand mesh reconstruction has been extensively investigated for years due to its wide range of applications in AR/VR \cite{bib:ARapp,bib:handAR}, behavior understanding \cite{bib:SocialAI,bib:Body2Hands}, \textit{etc}. Tremendous research efforts have been made towards this task, including \cite{bib:Ge19,bib:Zhou20,bib:YoutubeHand,bib:HIU}, to name a few.

The primary focus of typical existing methods is the reconstruction accuracy \cite{bib:Metro,bib:MeshGraphormer}, while real-world applications additionally necessitate inference efficiency and temporal consistency.
In particular, 3D hand information is a vital component in mobile applications \cite{bib:handAR}, where the devices comprise limited memory and computational budgets.
Thereby, this work aims to explore 3D hand reconstruction for mobile platforms. 

\begin{figure}[t]
\begin{center}
\includegraphics[width=\linewidth]{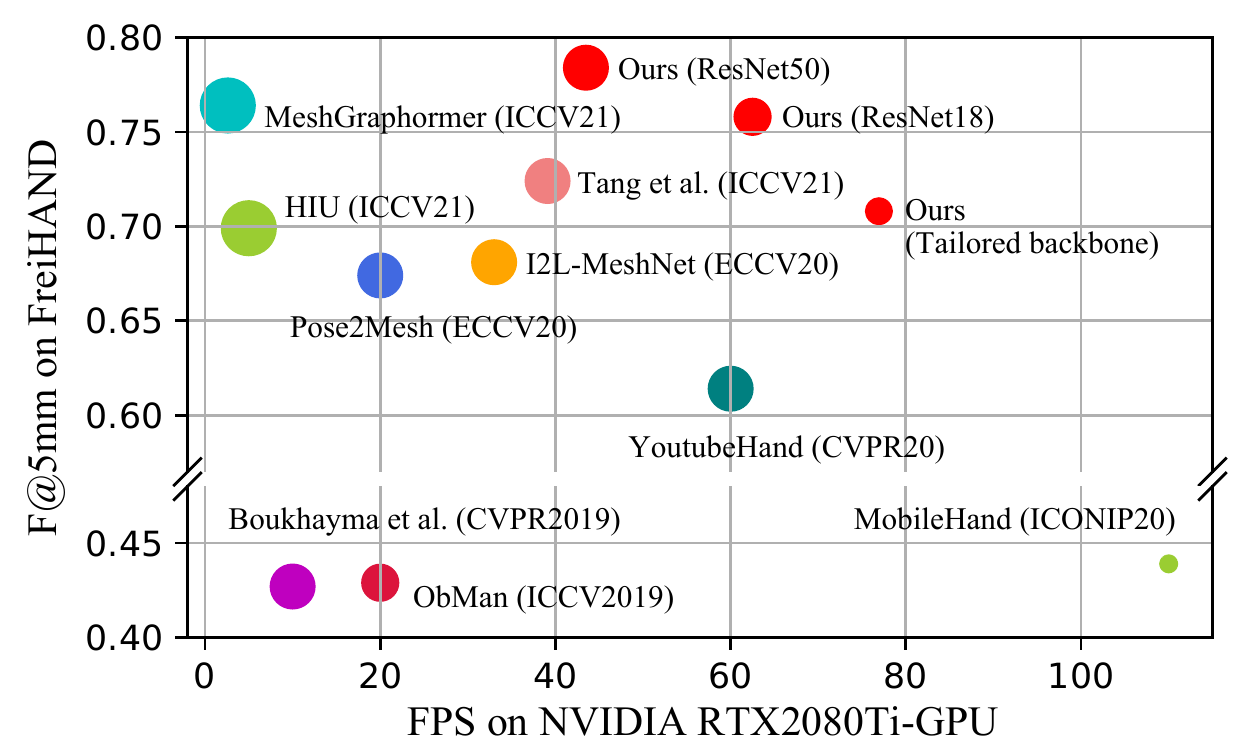}
\caption{Accuracy \textit{vs.} inference speed. The marker size is related to the model size. Besides, our tailored method can run at a fast speed on mobile CPUs.}
\label{fig:tradeoff}
\end{center}
\vspace{-0.55cm}
\end{figure}

A typical pipeline for single-view hand reconstruction includes three phases: 2D encoding, 2D-to-3D mapping, and 3D decoding.
In 2D encoding, existing approaches~\cite{bib:YoutubeHand,bib:CMR,bib:Metro,bib:MeshGraphormer} usually adopt computationally intensive networks \cite{bib:ResNet,bib:HRNet} to handle this highly non-linear task, which are hard to deploy on mobile devices.
Instead, if naively leveraging a mature mobile network (\textit{e.g.}, \cite{bib:MobileNetV3}) which is not tailored for our target task, the reconstruction accuracy dramatically degrades \cite{bib:MobileHand}.
Hence, our motivation is to develop a lightweight 2D encoding structure tailored to balance the inference efficiency and accuracy.
Besides, the efficiency of 2D-to-3D mapping and 3D decoding remains relatively unexplored. Thus, we intend to explore a lightweight yet effective lifting method to tackle the 2D-to-3D mapping problem and design an efficient graph operator for processing of 3D mesh data.

Although as crucial as accuracy in real-world applications, temporal coherence is usually neglected in the task of 3D hand reconstruction.
Previous methods \cite{bib:TCMR,bib:HMMR,bib:VIBE,bib:MEVA} adopt sequential models to incorporate both past and future semantic information for stable predictions.
Since this methodology is offline or computationally expensive, these approaches are difficult to use for mobile applications.
Hence, we are inspired to explore the temporal coherence with a non-sequential method.

In this work, we propose \textbf{Mob}ile Mesh \textbf{Recon}struction (MobRecon) for 3D hand to simultaneously explore superior accuracy, efficiency, and temporal coherence. 
For 2D encoding, we leverage the spirit of the hourglass network \cite{bib:Hourglass} to design efficient stacked encoding structures. 
As for 3D decoding, we propose a depth-separable spiral convolution (DSConv), which is a novel graph operator based on spiral sampling \cite{bib:SpiralConv}. The DSConv is inspired by depth-wise separable convolution \cite{bib:MobileNet}, leading to efficient handling of graph-structured mesh data.
Regarding the 2D-to-3D mapping, we propose a feature lifting module with map-based position regression (MapReg), pose pooling, and pose-to-vertex lifting (PVL) approaches. 
In this module, we first investigate the pros and cons of 2D pose representations based on heatmap or position regression, and then propose a hybrid method MapReg to simultaneously improve 2D pose accuracy and temporal consistency. 
Furthermore, the PVL transforms 2D pose encodings to 3D vertex features based on a learnable lifting matrix, resulting in enhanced 3D accuracy and temporal consistency. 
Compared to traditional approaches based on fully connected operation in a latent space \cite{bib:YoutubeHand,bib:Ge19,bib:CMR}, our feature lifting module also significantly reduces the model size.
In addition, we build a synthetic dataset with uniformly distributed hand poses and viewpoints.
Referring to Figure~\ref{fig:tradeoff}, we achieve better performance in terms of accuracy, speed, and model size.

Our main contributions are summarized as follows:
\begin{itemize}[leftmargin=*,topsep=3pt]
    \setlength\itemsep{-0.1em}
    \item We propose MobRecon as a mobile-friendly pipeline for hand mesh reconstruction, which only involves 123M multiply-add operations (Mult-Adds) and 5M parameters and can run up to 83FPS on Apple A14 CPU.
    \item We present lightweight stacked structures and DSConv for efficient 2D encoding and 3D decoding. 
    \item We propose a novel feature lifting module with MapReg, pose pooling, and PVL methods to bridge the 2D and 3D representations. 
    \item We demonstrate that our method achieves superior performance in terms of model efficiency, reconstruction accuracy, and temporal coherence via comprehensive evaluations and comparisons with state-of-the-art approaches. 
\end{itemize}

\section{Related Work}
\paragraph{Hand mesh estimation.}
Popular hand mesh estimation methods can be divided into five types, whose core ideas are based on the parametric model, voxel representation, implicit function, UV map, and vertex position, respectively. 

Model-based approaches \cite{bib:Zhang19,bib:Zhou20,bib:Bihand,bib:ObMan,bib:HIU,bib:Boukhayma19,bib:Baek19,bib:FreiHAND,bib:Chen21,bib:TravelNet,bib:Cao21,bib:Baek20,bib:Consist,bib:Liu21,bib:CPF,bib:Zhang21ICCV} typically use MANO \cite{bib:MANO} as the parametric model, which factorizes a hand mesh into coefficients of shape and pose. 
This pipeline, however, is not suitable for usage with a lightweight network because the coefficient estimation is a highly abstract problem that ignores spatial correlations \cite{bib:MobileHand}. 

Voxel-based approaches \cite{bib:Iqbal18,bib:I2L,bib:Semihand,bib:InterHand} describe the 3D data in a 2.5D manner. Moon \etal \cite{bib:I2L} proposed I2L-MeshNet, which divided the voxel space into three lixel spaces and used 1D heatmaps to reduce memory consumption. Despite this optimization, I2L-MeshNet still requires massive memory occupation to process the lixel-style 2.5D heatmaps. Hence, the voxel-based approach is not friendly for memory-constrained mobile devices.

The implicit function \cite{bib:OccNet} has merits of continuity and high resolution, which is recently used for digitizing articulated human \cite{bib:LEAP,bib:NASA,bib:LoopReg,bib:PIFu,bib:ARCH,bib:NeuralBody,bib:HALO,bib:GraspField}. 
However, the implicit methods usually need to compute thousands of 3D points, lacking efficiency in mobile settings.

Chen \etal \cite{bib:I2UV} treated hand mesh reconstruction as an image-to-image translation task, employing UV map to connect 2D and 3D spaces. This pipeline could be improved by incorporating geometry correlations.

Vertex-based methods \cite{bib:Ge19,bib:YoutubeHand,bib:CMR,bib:Metro,bib:MeshGraphormer} predict 3D vertex coordinates directly, which usually follow a procedure of 2D encoding, 2D-to-3D mapping, and 3D decoding. For example, Kulon \etal \cite{bib:YoutubeHand} designed an encoder-decoder based on ResNet \cite{bib:ResNet}, global pooling, and spiral convolution (SpiralConv) \cite{bib:SpiralConv} to obtain 3D vertex coordinates. We re-construct the vertex-based pipeline with efficient modules, \emph{i.e.}, lightweight stacked structures for 2D encoding, feature lifting module for 2D-to-3D mapping, and DSConv for 3D decoding. As a result, we achieve high reconstruction accuracy and across-time consistency.

\begin{figure*}[t]
\begin{center}
\includegraphics[width=\linewidth]{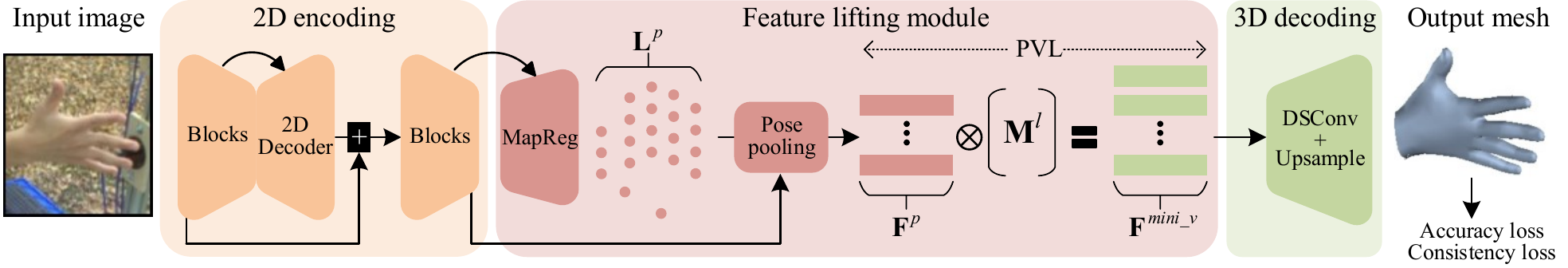}
\caption{Overview of our MobRecon framework.}
\label{fig:arch}
\end{center}
\vspace{-0.55cm}
\end{figure*}

\vspace{-0.4cm}
\paragraph{Lightweight networks.}
For a timely fashion on a computationally limited platform, lightweight networks have been studied for years such as \cite{bib:MobileNet,bib:MobileNetV2,bib:MobileNetV3,bib:GhostNet}. 
We leverage popular efficient ideas to design stacked networks for Euclidean 2D images and a graph network for non-Euclidean 3D meshes. More specifically, we propose a feature lifting module to deal with the problem of 2D-to-3D mapping efficiently. As a highly related work, MobileHand \cite{bib:MobileHand} was able to run at 75FPS on a mobile CPU. In contrast, our MobRecon achieves more powerful performance on accuracy and inference speed (as shown in Table~\ref{tab:freihand}).

\vspace{-0.4cm}
\paragraph{Temporally coherent human reconstruction.}
There has been limited research on the temporal coherence of human/hand mesh reconstruction, yet it is as crucial as reconstruction accuracy in real-world applications.
Previous research \cite{bib:TCMR,bib:HMMR,bib:VIBE,bib:MEVA} has concentrated on temporal performance with temporal approaches.
Kocabas \etal \cite{bib:VIBE} used a bi-directional gated recurrent unit \cite{bib:GRU} to fuse across-time features so that SMPL \cite{bib:SMPL} parameters could be regressed with temporal cues. 
This sequential manner could raise computational costs and even requires future information \cite{bib:TCMR}. In contrast, we design a feature lifting module with MapReg to enhance the temporal coherence for a non-sequential single-view method.

\vspace{-0.4cm}
\paragraph{Pixel-aligned representation.} Convolutional features are made up of dense and regularly structured 2D cues, but sparse and unordered points represent 3D data. To better extract image features to describe 3D information, recent works usually adopt pixel-aligned representations \cite{bib:PIFu,bib:Pose2Pose,bib:HoloPose,bib:Zhang21ICCV,bib:HandsFormer,bib:PAMIR}. 
Inspired by them, we use the idea of pixel alignment for feature lifting and design PVL to transform pose-aligned encodings to vertex features.

It is noteworthy that the above-mentioned literature used heatmap \cite{bib:HoloPose,bib:HandsFormer,bib:Zhang21ICCV} or position \cite{bib:PIFu,bib:PAMIR} as the 2D representations. Li \etal \cite{bib:RLE} analyzed the heatmap- and position-based human pose in terms of accuracy and proposed RLE to achieve high-accuracy regression. We consider these two representations from the view of temporal coherence and propose MapReg to integrate the merits of heatmap- and position-based 2D representations. Compared to \cite{bib:RLE}, we have different insights and study perspectives, so RLE and our MapReg could complement each other.


\section{Our Method}
With a single-view image as the input, we aim to infer 3D hand mesh with predicted vertices $\mathbf V=\{\mathbf v_i\}_{i=1}^V$ and pre-defined faces $\mathbf C=\{\mathbf{c}_i\}_{i=1}^C$. 
Figure~\ref{fig:arch} illustrates the overall architecture of MobRecon that includes three phases. For 2D encoding, we leverage convolutional networks to extract image features. In the feature lifting module, 2D pose is delineated with MapReg, followed by pose pooling to retrieve pose-aligned features. Then, vertex features are obtained with PVL. For 3D decoding, we develop an efficient graph network to predict vertex coordinates.

\subsection{Stacked Encoding Network}

\begin{figure}[t]
\begin{center}
\includegraphics[width=\linewidth]{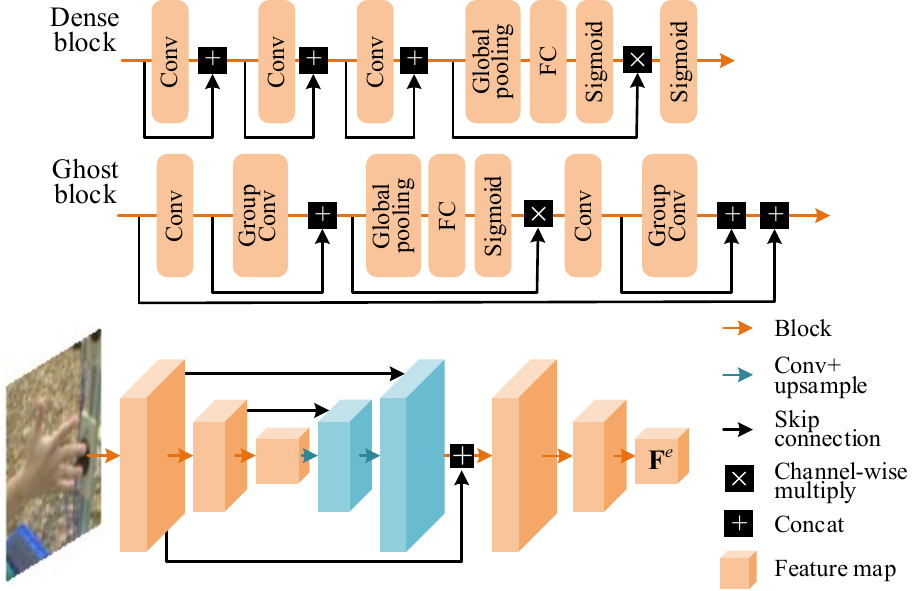}
\caption{Our tailored stacked encoding structure.}
\label{fig:stack}
\end{center}
\vspace{-0.7cm}
\end{figure}

Inspired by the hourglass network \cite{bib:Hourglass}, we develop a stacked encoding network to obtain gradually refined encoding features. As shown in Figure~\ref{fig:stack}, the stacked network consists of two groups of cascaded encoding blocks, the first of which is followed by an upsampling module for feature fusion. With a single-view image as the input, the encoding feature $\mathbf F^e\in \mathbb R^{C^e\times H^e\times W^e}$ is generated, where $C, H, W$ denote tensor channel size, height, and width.

We design two block alternatives. According to DenseNet \cite{bib:DenseNet} and SENet \cite{bib:SENet}, we present a dense block to form DenseStack (see Figure~\ref{fig:stack}). To further reduce model size, we leverage the ghost operation \cite{bib:GhostNet} to develop GhostStack, where cheap operations can produce ghost features based on primary features. With $128\times 128$ input resolution, the DenseStack involves $373.0$M Mult-Adds and $6.6$M parameters while the GhostStack consists of $96.2$M Mult-Adds and $5.0$M parameters. In contrast, a stacked network with ResNet18 has $2391.3$M Mult-Adds and $25.2$M parameters, prohibiting it from mobile applications.

\subsection{Feature Lifting Module}

\begin{figure}[t]
\begin{center}
\includegraphics[width=\linewidth]{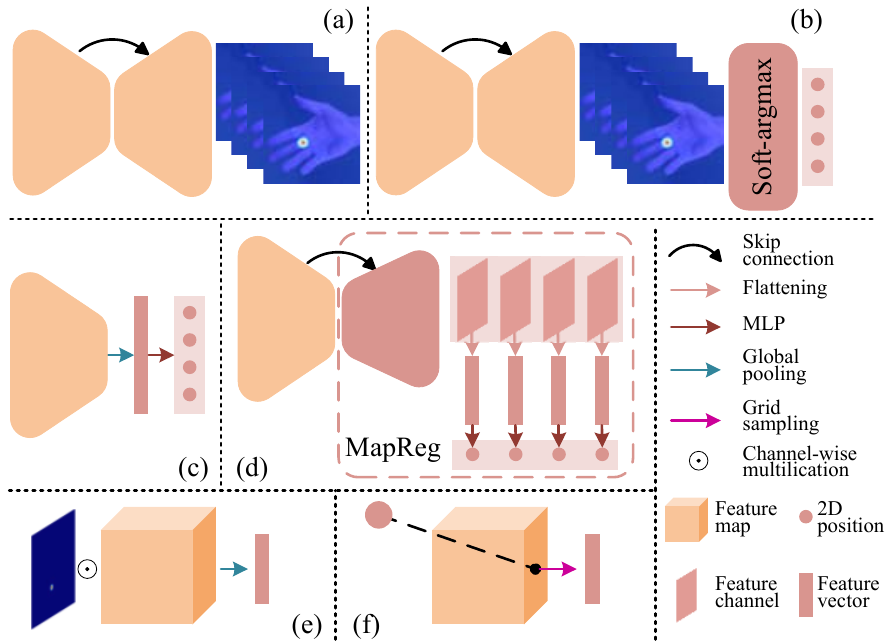}
\caption{Comparisons of 2D representations and pose pooling methods. (a) heatmap, (b) heatmap + soft-argmax, (c) regression-based position, (d) MapReg-based position, (e) joint-wise pooling with a heatmap, (f) grid sampling with a 2D position. For better visualization, only 4 landmarks are illustrated.}
\label{fig:rep2d}
\end{center}
\vspace{-0.55cm}
\end{figure}

\textit{Lifting} means 2D-to-3D mapping \cite{bib:Bihand}. 
For feature lifting, two problems should be concerned: (1) how to collect 2D features and (2) how to map them to 3D domain. To this end, previous methods \cite{bib:YoutubeHand,bib:Ge19,bib:CMR} tend to embed $\mathbf F^e$ as a latent vector via the global average pooling operation. Then, the latent vector is mapped to 3D domain with a fully connected layer (FC), and vertex features are obtained with vector re-arrangement. This manner brings increased model size due to the large dimension of FC, \textit{i.e.}, this layer contains 3.2M parameters when $C^e$=256. 

Recent researches report pixel-aligned feature extraction based on 2D landmarks and pixel-aligned feature pooling \cite{bib:PIFu,bib:HoloPose,bib:HandsFormer,bib:Zhang21ICCV,bib:PAMIR}. Heatmap $\mathbf H^p$ is usually employed to encode 2D landmarks \cite{bib:HoloPose,bib:Ge19,bib:CMR,bib:HIU}, which derives more accurate landmarks compared with direct regression of the 2D positions $\mathbf L^p$ \cite{bib:RLE,bib:DeepPose,bib:Carreira16,bib:Sun17,bib:Wei20}. 

\vspace{-0.4cm}
\paragraph{Map-based position regression.}
Figure~\ref{fig:rep2d} comprehensively investigates these 2D representations. It is seen that $\mathbf H^p$ is a high-resolution representation (Figure~\ref{fig:rep2d}(a)). As shown in Figure~\ref{fig:rep2d}(b), soft-argmax \cite{bib:softargmax} is a differentiable technique that 
decodes $\mathbf H^p$ to the corresponding 2D position.
In contrast, direct position regression (Figure~\ref{fig:rep2d}(c)) is a low-resolution representation, which can yield $\mathbf L^p$ without resorting to $\mathbf H^p$. 
It has been proven that the human pose estimation task requires both low- and high-resolution encodings \cite{bib:HRNet}. Hence, the skip connection that fuses features is the primary cause behind the superior accuracy of $\mathbf H^p$ \cite{bib:Hourglass}. However, our critical insight is that the global feature (with resolution of $1\times 1$) used to predict $\mathbf L^p$ (see Figure~\ref{fig:rep2d}(c)) 
shall induce better temporal coherence because of global semantics and receptive field. This global property can better describe articulated relation of hand pose. Conversely, $\mathbf H^p$ is predicted with convolution and high-resolution features, where limited receptive field leads to the lack of inter-landmark constraints.

As shown in Figure~\ref{fig:rep2d}(d), we propose a middle-resolution method MapReg to combine the advantages of heatmap- and position-based paradigms by (1) fusing low- and high-resolution features for accuracy and (2) using global receptive field for temporal coherence. To this end, we incorporate the skip connection in the position regression paradigm, producing a spatially small-size (\eg $16\times 16$) feature map. Each feature channel is then flattened into a vector, followed by a multi-layer perceptron (MLP) to generate a 2D position. In this manner, we obtain middle-resolution spatial complexity which is superior than heatmap since only two 2$\times$-upsampling operations are involved. 

\vspace{-0.4cm}
\paragraph{Pose pooling.}
After obtaining 2D representations, pixel-aligned features can be retrieved. We call this process pose pooling and capture pose-aligned features with $N$ 2D joint landmarks. If heatmap $\mathbf H^p$ is employed as the 2D representation, the pose pooling is conducted with joint-wise pooling \cite{bib:Zhang21ICCV} (Figure~\ref{fig:rep2d}(e)), which can be given as
\vspace{-0.2cm}
\begin{equation}
    \mathbf F^p = [\Psi(\mathbf F^e \odot {\rm interpolation}(\mathbf H^p_i))]_{i=1,2,...,N},
\vspace{-0.2cm}
\end{equation}
where $[\cdot]$ denotes concatenation. First, the spatial size of $\mathbf H^p$ is interpolated to $H^e\times W^e$, and then channel-wise multiplication is adopted between interpolated $\mathbf H^p$ and $\mathbf F^e$. In this manner, features unrelated to joint landmarks are suppressed. For extracting joint-wise features, feature reduction $\Psi$ is designed. In detail, $\Psi$ denotes global max-pooling or spatial sum reduction, which produces a $C^e$-length feature vector. After concatenation, $\mathbf F^p\in\mathbb R^{N\times C^e}$ indicates the pose-aligned feature . 

If we use $\mathbf L^p$ instead of $\mathbf H^p$ to describe 2D pose, pose pooling can be achieved with grid sampling \cite{bib:PIFu} (Figure~\ref{fig:rep2d}(f)) as follows,
\vspace{-0.2cm}
\begin{equation}
    \mathbf F^p = [\mathbf F^e(\mathbf L^p_i)]_{i=1,2,...,N}.
\vspace{-0.2cm}
\end{equation}

As a result, the convolutional encoding $\mathbf F^e$ is transformed to the pose-aligned representation $\mathbf F^p$.

\vspace{-0.4cm}
\paragraph{Pose-to-vertex lifting.}
Referring to Figure~\ref{fig:arch}, we design a linear operator for feature mapping towards 3D space with a few learnable parameters, namely PVL. MANO-style hand mesh \cite{bib:MANO} comprises $V$ vertices and $N$ joints, where $V=778,N=21$. Because $V\gg N$, it is hard to transform $\mathbf F^p$ to $V$ vertex features. Instead, we downsample the template mesh 4-times by a factor of 2 \cite{bib:Sample}, and obtain a minimal-size hand mesh with $V^{mini}=49$ vertices. Then, we design a learnable lifting matrix $\mathbf M^l\in\mathbb R^{V^{mini}\times N}$ for 2D-to-3D feature mapping. Thereby, PVL is given as
\vspace{-0.2cm}
\begin{equation}
    \mathbf F^{mini\_v} = \mathbf M^l \cdot \mathbf F^{p},
    \vspace{-0.2cm}
\end{equation}
where $\mathbf F^{mini\_v}$ denotes minimal-size vertex feature. The PVL reduces the computational complexity of feautre mapping  from $\mathcal O(V^{mini}C^{e2})$ \cite{bib:YoutubeHand,bib:CMR,bib:Ge19} to $\mathcal O(NV^{mini}C^e)$.

\subsection{Depth-Separable SpiralConv}

As a graph operator, SpiralConv \cite{bib:SpiralConv} is equivalent to the Euclidean convolution, which designs a spiral neighbor as
\begin{equation}
\label{equ:spiral}
\begin{aligned}
    0\text{-ring}(\mathbf v) & =\{\mathbf v\} \\
    (k+1)\text{-ring}(\mathbf v) & =\mathbb N(k\text{-ring}(\mathbf v)) \setminus k\text{-disk}(\mathbf v) \\
    k\text{-disk}(\mathbf v) & =\cup_{i=0,..,k}i\text{-ring}(\mathbf v), \\
\end{aligned}
\end{equation}
where $\mathbb N$ extracts the neighbourhood of a vertex $\mathbf v$. With $k\text{-disk}(\mathbf v)$, SpiralConv formulates convolution as a sequential problem and leverages LSTM \cite{bib:LSTM} for feature fusion:
\begin{equation}
\label{equ:lstm}
\mathbf f^{out}_{\mathbf v} = {\rm LSTM}(\mathbf f_{\mathbf v'}), \quad \mathbf v' \in k\text{-disk}(\mathbf v),
\end{equation}
where $\mathbf f_{\mathbf v}\in \mathbb R^{1 \times D}$ denotes the feature vector at $\mathbf v$ with dimension $D$. SpiralConv with LSTM could be potentially slow because of serial sequence processing.

By explicitly formulating the order of aggregating neighboring vertices, SpiralConv++~\cite{bib:Spiral++} presents an efficient version of SpiralConv. For high efficiency, SpiralConv++ only adopts a fixed-size spiral neighbor with $S$ vertices and leverages FC to fuse these features:
\begin{equation}
\label{equ:spiral++}
    \mathbf f^{out}_{\mathbf v} = \mathbf W \cdot [\mathbf f_{\mathbf v'}] + \mathbf b, \quad {\mathbf v'} \in k\text{-disk}({\mathbf v})_S,
\vspace{-0.2cm}
\end{equation}
where $[\cdot]$ denotes concatenation; $k\text{-disk}({\mathbf v})_S$ contains the first $S$ elements in $k\text{-disk}({\mathbf v})$; $\mathbf W$ and $\mathbf b$ are learnable parameters. SpiralConv++ significantly increases model size because of the large dimension of FC.

As an efficient graph convolution, we propose DSConv with the spirit of depth-wise separable convolution \cite{bib:MobileNet}. For a vertex ${\mathbf v}$, $k\text{-disk}({\mathbf v})_S$ is sampled following Equation \ref{equ:spiral}. DSConv comprises a depth-wise operation and a point-wise operation, the former of which can be formulated as
\begin{equation}
\label{equ:dw}
    \mathbf f^d_{\mathbf v} = [\mathbf W^d_i \cdot [\mathbf f_{\mathbf v',i}]]_{i=1}^D, \quad {\mathbf v'} \in k\text{-disk}({\mathbf v})_S.
\vspace{-0.2cm}
\end{equation}

\begin{figure}[t]
\begin{center}
\includegraphics[width=\linewidth]{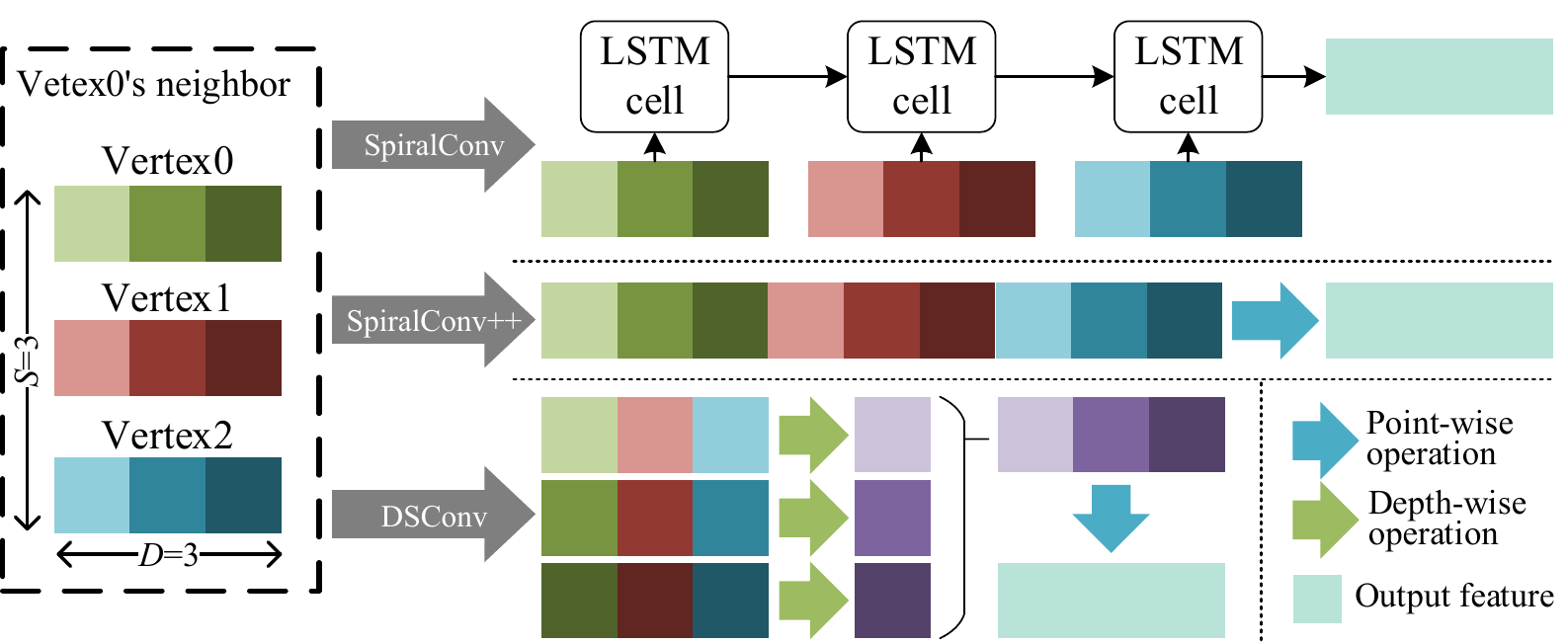}
\caption{Comparison of SpiralConv, SpiralConv++, and DSConv. For better visualization, a case with $S=D=3$ is shown.}
\label{fig:dsconv}
\end{center}
\vspace{-0.55cm}
\end{figure}

Then, point-wise operation can be formulated as
\begin{equation}
\label{equ:pw}
    \mathbf f^{out}_{\mathbf v} = \mathbf W^p \cdot \mathbf f^d_{\mathbf v}.
\vspace{-0.2cm}
\end{equation}

Figure~\ref{fig:dsconv} illustrates the difference among SpiralConv, SpiralConv++, and our DSConv. The SpiralConv++'s computational complexity obeys $\mathcal O(SD^2)$, whereas that of DSConv is $\mathcal O(SD+D^2)$. Hence, we essentially improve the efficiency with the separable structure.

The 3D decoder is built with four blocks, each of which involves upsampling, DSConv, and ReLU. In each block, vertex features are upsampled by a factor of 2 and then processed by DSConv. Finally, vertex coordinates $\mathbf V$ are predicted by a DSConv.

\subsection{Loss Functions}

\paragraph{Accuracy loss.}
We use $L_1$ norm for the 3D mesh loss $\lossmesh$ and 2D pose loss $\lossposetwo$.
Normal loss $\lossnormal$ and edge length loss $\lossedge$ are adopted for mesh smoothness according to \cite{bib:Ge19}. Formally, we have
\begin{equation}
\begin{aligned}
    & \lossmesh = ||\mathbf V-\mathbf V^\star||_1, \lossposetwo = ||\mathbf L^p-\mathbf L^{p,\star}||_1 \\
    & \lossnormal =\sum_{\mathbf{c}\in \mathbf C}\sum_{(i,j)\subset\mathbf{c}}|\frac{\mathbf V_i-\mathbf V_j}{||\mathbf V_i-\mathbf V_j||_2}\cdot \mathbf n_\mathbf{c}^\star| \\
    & \lossedge =\sum_{\mathbf{c}\in \mathbf C}\sum_{(i,j)\subset\mathbf{c}} |||\mathbf V_i-\mathbf V_j||_2 - ||\mathbf V^\star_i-\mathbf V^\star_j||_2|,
\label{eqn:loss}
\end{aligned}
\end{equation}
where $\mathbf C, \mathbf V$ are face and vertex sets of a mesh; $\mathbf n^\star_\mathbf{c}$ indicates unit normal vector of face $\mathbf{c}$; $\star$ denotes the ground truth.

\vspace{-0.4cm}
\paragraph{Consistency loss.}


Inspired by the self-supervision task \cite{bib:Spurr21,bib:HandCo}, we design the consistency supervision based on augmentation without the need of temporal data. 
That is, two views can be derived from an image sample based on 2D affine transformation and color jitter. We denote relative affine transformation between two views as $T_{1\rightarrow2}$, which contains relative rotation $R_{1\rightarrow2}$. Similar to \cite{bib:DenseUNet,bib:Semihand}, we conduct consistency supervision in both 3D and 2D space:
\begin{equation}
\begin{aligned}
    & \lossconthree = ||R_{1\rightarrow2}\mathbf{V}_{\text{view1}}-\mathbf{V}_{\text{view2}}||_1 \\
    & \losscontwo = ||T_{1\rightarrow2}\mathbf{L}^p_{\text{view1}}-\mathbf{L}^p_{\text{view2}}||_1. \\
\end{aligned}
\end{equation}
Although $T$ contains the variations of rotation, shift, and scale in 2D space, only $R$ affects $\mathbf{V}$ because it is rooted by the wrist landmark.

Our overall loss function is $\losstotal=\lossmesh+\lossposetwo+\lossnormal+\lossedge+\losscontwo+\lossconthree$.

\section{Experiments}
\label{sec:experiments}

\subsection{Implementation Details}
We use the Adam optimizer \cite{bib:Adam} to train the network with a mini-batch size of $32$. All models in our experiments are trained for $38$ epochs. The initial learning rate is $10^{-3}$, which is divided by $10$ at the $30$th epoch. As for hyperparameters, the input resolution is $128\times 128$, and $S=9,C^e=256,D=\{256, 128, 64, 32\}$.

Existing datasets for 3D hand pose estimation (\eg \cite{bib:FreiHAND,bib:YoutubeHand,bib:HO3D}) usually suffer from long-tailed distribution of hand pose and viewpoint. Hence, we develop a synthetic dataset with 1520 poses and 216 viewpoints, both of which are uniformly distributed in their respective spaces. Because of this superior property, it can serve as a good complement during training. We redirect the readers to supplementary material for details.

\subsection{Evaluation Criterion}
We conduct experiments on several commonly-used benchmarks as listed below.
\begin{description}[leftmargin=10pt,itemsep=0pt,parsep=0pt,noitemsep,topsep=3pt]
\item[FreiHAND]\cite{bib:FreiHAND} 
contains 130,240 training images and 3,960 evaluation samples. The annotations of the evaluation set are not available, so we submit our predictions to the official server for online evaluation.
\item[Rendered Hand Pose Dataset (RHD)]\cite{bib:RHD}
consists of 41,258 and 2,728 synthetic hand data for training and testing on hand pose estimation, respectively.
\item[HO3Dv2]\cite{bib:HO3D}
is a 3D hand-object dataset that contains 66,034 training samples and 11,524 evaluation samples. The annotations of the evaluation set are not available, so we use the official server for online evaluation. We also use this dataset to evaluate temporal performance.
\end{description}

We use the following metrics in quantitative evaluations.
\begin{description}[leftmargin=10pt,itemsep=0pt,parsep=0pt,noitemsep,topsep=3pt]
\item[MPJPE/MPVPE] 
measures the mean per joint/vertex position error by Euclidean distance (mm) between the estimated and ground-truth coordinates.
\item[PA-MPJPE/MPVPE] 
is a modification of MPJPE/MPVPE with Procrustes analysis~\cite{bib:PA}, ignoring global variation. For conciseness, this metric is abbreviated as PJ/PV.
\item[Acc] captures the acceleration of 2D/3D joint landmarks in $\mathrm{pixel/s^2}$ or $\mathrm{mm/s^2}$ to reflect temporal coherence.
\item[AUC] 
means the area under the curve of PCK (percentage of correct keypoints) \emph{vs}. error thresholds of $n\sim$50mm (for 3D measurement) or 0$\sim$30 pixel (for 2D measurement). According to \cite{bib:FreiHAND,bib:RHD}, $n=0$ or $20$.
\item[F-Score] 
is the harmonic mean between recall and precision between two meshes \emph{w.r.t.} a specific distance threshold. F@5/F@15 corresponds to a threshold of 5mm/15mm.
\item[Mult-Adds] counts multiply-add operations.
\item[\#Param] indicates the number of parameters. 
\end{description}

\begin{table}[t]
\small
\renewcommand{\arraystretch}{1.}
\setlength\tabcolsep{2pt}
\centering
\begin{tabular}{c c c c c c}
\hline
2D encoding   & Mult-Adds & \#Param & Fine-tuning data & PJ$\downarrow$ \\

\hline

\multicolumn{6}{c}{ \small{\textit{From-scratch fine-tuning}} } \\
ResNet18-Stack  & 2391.3M & 25.2M & FreiHAND  & 8.86 \\
MobileNet-Stack  & 100.2M  & 1.7M & FreiHAND & 14.20 \\
GhostStack & 96.2M  & 5.0M & FreiHAND & 13.09 \\
DenseStack & 373.0M & 6.6M & FreiHAND & 9.56  \\

\hline
\multicolumn{6}{c}{ \small{\textit{Pre-training w/ ImageNet}} } \\
ResNet18-Stack  & 2391.3M & 25.2M & FreiHAND  & 8.21 \\
MobileNet-Stack  & 100.2M  & 1.7M & FreiHAND & 13.68 \\

\hline
\multicolumn{6}{c}{ \small{\textit{Pre-training w/ ours}} } \\
MobileNet-Stack  & 100.2M  & 1.7M & FreiHAND & 12.45 \\
\cdashline{1-6}[0.8pt/2pt]
\multirow{2}*{GhostStack} & \multirow{2}*{96.2M}  & \multirow{2}*{5.0M} & FreiHAND & 10.05  \\
 & & & FreiHAND+ours & 8.89 \\
\cdashline{1-6}[0.8pt/2pt]
\multirow{2}*{DenseStack} & \multirow{2}*{373.0M} & \multirow{2}*{6.6M} & FreiHAND & 7.77  \\
 & & & FreiHAND+ours & \bf 7.55  \\
\hline
\end{tabular}
\caption{Ablation studies of 2D encoding methods and our complement data on FreiHAND. Mult-Adds and \#Param are \textit{w.r.t.} the 2D encoding network.}
\label{tab:stack}
\vspace{-0.2cm}
\end{table}

\begin{figure}[t]
\begin{center}
\includegraphics[width=0.9\linewidth]{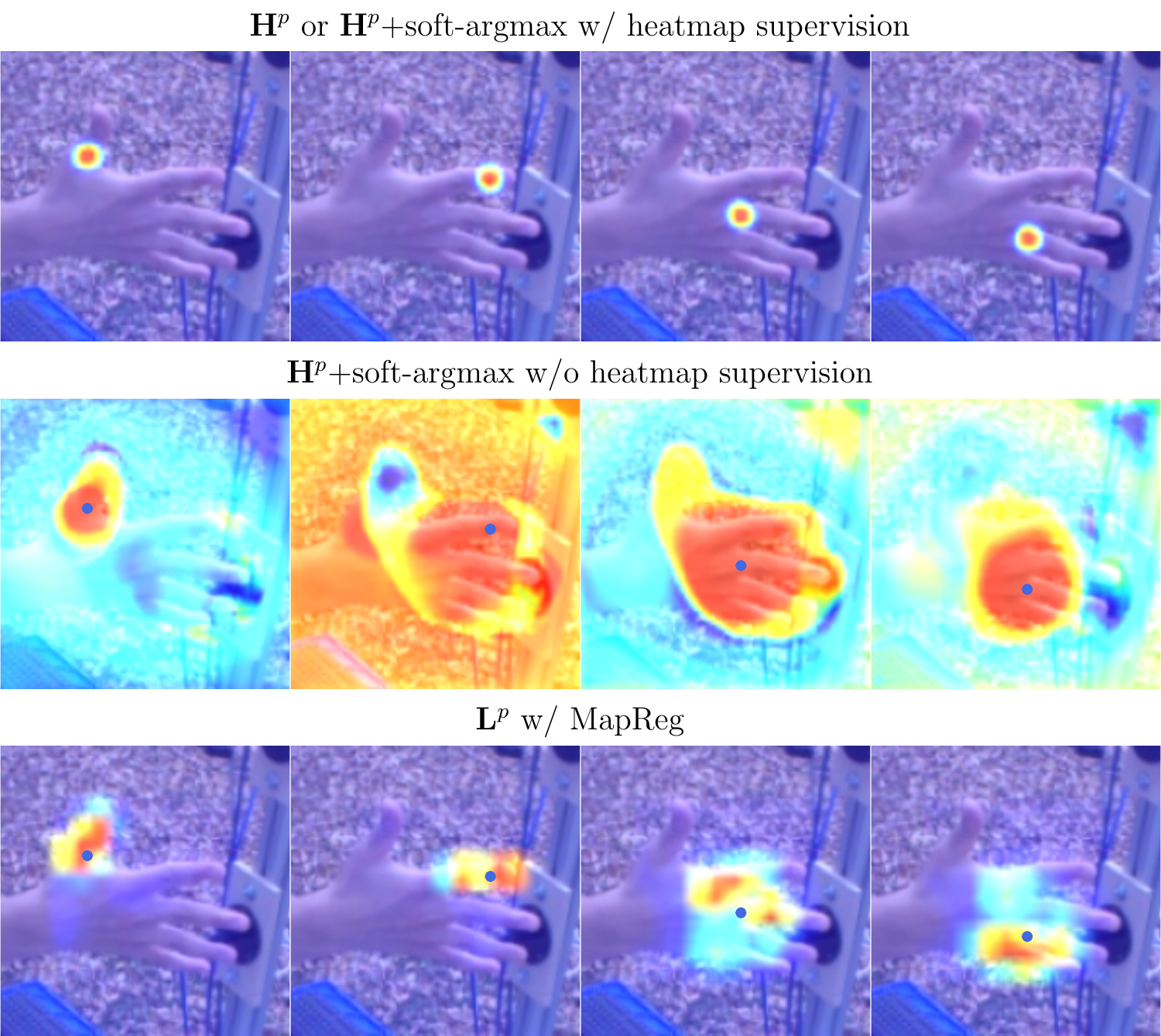}
\caption{Visualization of \textit{maps} in various 2D pose representations. Different from heatmap that focus on individual landmark, the \textit{map} in MapReg adaptively describes inter-landmark constraints.}
\label{fig:map}
\end{center}
\vspace{-0.4cm}
\end{figure}

\begin{figure}[t]
\begin{center}
\includegraphics[width=\linewidth]{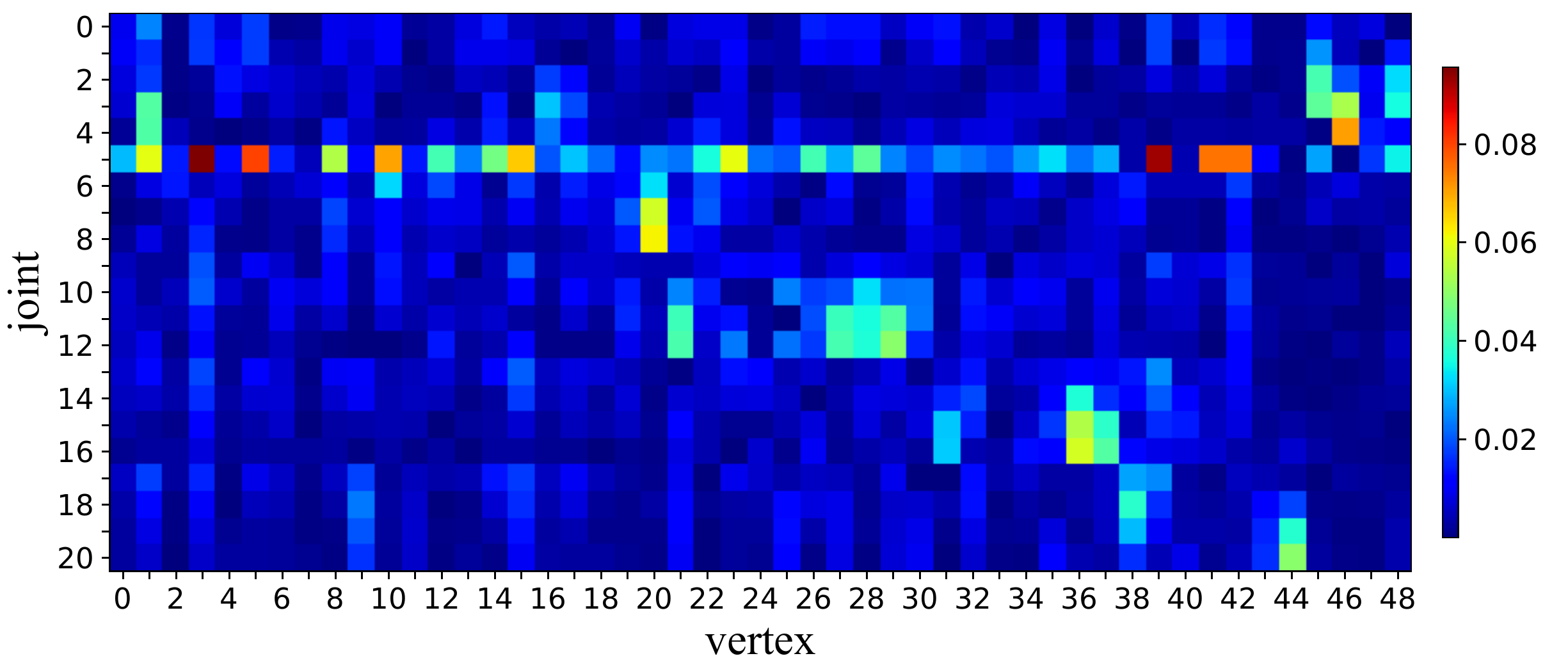}
\caption{Learned lifting matrix.}
\label{fig:matrix}
\end{center}
\vspace{-0.4cm}
\end{figure}

\begin{figure}[t]
\begin{center}
\includegraphics[width=0.8\linewidth]{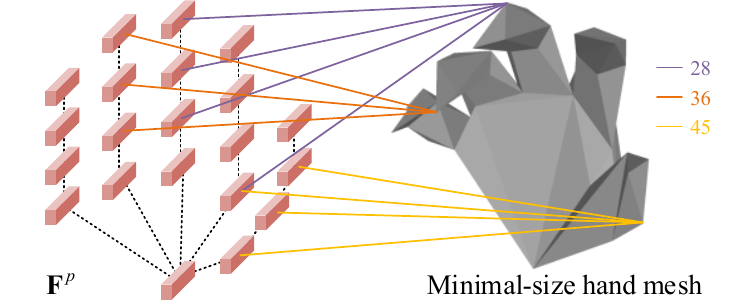}
\caption{Highly relevant connections for several vertices in the  lifting matrix. The numbers are vertex indices.}
\label{fig:lifting}
\end{center}
\vspace{-0.6cm}
\end{figure}

\begin{table*}[t]
\small
\renewcommand{\arraystretch}{1.}
\setlength\tabcolsep{1.5pt}
\centering
\begin{tabular}{c | c c c c | c c c c c c c}
\hline
\multirow{2}{*}{2D representation}   & \multicolumn{4}{c|}{2D part}  & \multicolumn{7}{c}{2D-to-3D part} \\
& Mult-Adds & \#Param & 2D AUC$\uparrow$ & 2D Acc$\downarrow$ & Pose pooling & Lifting & Mult-Adds & \#Param & PJ$\downarrow$ & 3D AUC$\uparrow$ & 3D Acc $\downarrow$  \\
\hline
$-$   & \multicolumn{4}{c|}{$-$} & Global pooling & FC & 3.2M & 3.2M & 7.55 & 0.850 & 10.12 \\
$\mathbf H_p$ w/o skip.& 57.7M & 1.4M & 0.796 & 2.57 & \multicolumn{7}{c}{$-$} \\
$\mathbf H_p$          & 57.7M & 1.4M & 0.857 & 2.41 & Joint-wise pooling & PVL & 0.4M & 1029 & 7.41 & 0.852 & 9.11 \\
$\mathbf H_p$+soft-argmax      & 57.7M & 1.4M & 0.865 & 2.32 & Grid sampling  & PVL & 0.3M & 1029 & 7.28 & 0.856 & 8.32 \\
$\mathbf L_p$ w/ reg.         & 0.1M & 0.1M & 0.847 & 2.16 & Grid sampling  & PVL & 0.3M & 1029 & 7.11 & 0.859 & 6.14 \\
$\mathbf L_p$ w/ MapReg (ours)& 47.5M & 1.4M & \bf 0.870 & \bf 2.04 & Grid sampling  & PVL & 0.3M & 1029 & \bf 6.95 & \bf 0.862 & \bf 5.41 \\
\hline
\end{tabular}
\caption{Ablation study of feature lifting module. The first row is the setting of CMR \cite{bib:CMR} (\ie the last row of Table~\ref{tab:stack}); skip. and reg. denote skip connection and direct regression; 2D and 3D accuracy are tested on RHD and FreiHAND, respectively; Acc is \textit{w.r.t.} HO3Dv2.}
\label{tab:2d}
\vspace{-0.3cm}
\end{table*}

\begin{table}[t]
\small
\renewcommand{\arraystretch}{1.}
\setlength\tabcolsep{1.5pt}
\centering
\begin{tabular}{c c | c c c c}
\hline
$\lossconthree$ & $\losscontwo$ & PJ $\downarrow$ & 3D AUC$\uparrow$ & 2D Acc$\downarrow$ & 3D Acc$\downarrow$ \\
\hline
             &              & 6.95 & 0.862 & 2.04 & 5.41 \\
$\checkmark$ &              & \bf 6.85 & \bf 0.864 & 2.03 & 4.78 \\
$\checkmark$ & $\checkmark$ & \bf 6.85 & \bf 0.864 & \bf 1.98 & \bf 4.75 \\
\hline
\end{tabular}
\caption{Ablation study of consistency learning. The accuracy is \textit{w.r.t.} FreiHAND and Acc is \textit{w.r.t.} HO3Dv2.}
\label{tab:consist}
\vspace{-0.3cm}
\end{table}

\subsection{Ablation studies}

\paragraph{Our stacked networks and complement data.}

We use CMR \cite{bib:CMR} as the baseline, which adopts FC for 2D-to-3D feature mapping and SpiralConv++ for 3D decoding. With the same hyperparameters, we only change the 2D encoding network and data setting for comparisons. On the one hand, we study different stacked structures with from-scratch training. As shown in Table~\ref{tab:stack}, we use ResNet \cite{bib:ResNet} and MobileNet \cite{bib:MobileNetV2} to design stacked structures, the former of which contains intensive computation costs. Although MobileNet is computational tractability, it whittles down the performance (\textit{i.e.}, PAMPJPE) by a large margin. On the other hand, ImageNet \cite{bib:ImageNet} is used for classification task, the knowledge from which is hard to transfer towards 2D/3D position regression. Consequently, the ImageNet pre-training brings less than 1mm PAMPJPE improvement.   

Referring to Table~\ref{tab:stack}, our complement data can be included in both pre-training and fine-tuning steps, and DenseStack/GhostStack induces 7.55/8.89mm PAMPJPE.
Hence, we significantly reduce the computational cost without sacrificing reconstruction accuracy with our tailored encoding structures and dataset, making our DenseStack/GhostStack suitable for mobile environments.

\vspace{-0.4cm}
\paragraph{Feature lifting module.}

With DenseStack, we perform a joint study of accuracy and temporal coherence. In Table~\ref{tab:2d}, we do not use sequential module, temporal optimization, or post-processing. First, we explore various 2D pose representations in detail. $\mathbf H^p$ (Figure~\ref{fig:rep2d}(a)) is a high-accuracy representation, and the skip connection proves critical to fuse high- and low-resolution features. Hence, when the skip connection is removed, $\mathbf H^p$ performs poorly in accuracy. As a differentiable form of picking maximum position, soft-argmax can produce smoother 2D positions from $\mathbf H^p$, hence improving both 2D AUC and Acc. In addition, $\mathbf L^p$ w/ reg. (Figure~\ref{fig:rep2d}(c)) has relatively modest accuracy but produces better temporal performance because of global receptive field. Ultimately, our MagReg achieves better 2D accuracy and temporal coherence by integrating the merits of heatmap- and regression-based paradigms.

To clearly reveal details, we illustrate \textit{maps} in Figure~\ref{fig:map}. As for $\mathbf H^p$+soft-argmax, we additionally use the same loss setting (Equation \ref{eqn:loss}) for training, \textit{i.e.}, heatmap supervision is not involved. This manner naively induces a smooth version of $\mathbf H^p$ because the heuristic soft-argmax neglects visual semantics. In terms of MapReg, we present the \textit{map} before it is flattened into a vector. Different from $\mathbf H^p$, the \textit{map} can adaptively describe joint landmark constraints, and then the 2D positions are predicted using adaptive local-global information. For example, the entire thumb is activated when predicting a landmark on it. 
Hence, our MapReg can produce more reasonable articulated structures (as shown in the supplemental material) to improve temporal coherence.

Subsequently, we explore 3D performance with 2D pose-aligned features. 
During pose pooling, joint-wise pooling \cite{bib:Zhang21ICCV} (Figure~\ref{fig:rep2d}(e)) can be used to obtain 2D pose-aligned features based on $\mathbf H^p$, while grid sampling \cite{bib:PIFu} (Figure~\ref{fig:rep2d}(f)) is commonly adopted when soft-argmax or $\mathbf L^p$ is utilized. 
It is noteworthy that although $\mathbf L^p$ w/ reg. lags behind in terms of 2D accuracy, it produces better PAMPJPE than $\mathbf H^p$+soft-argmax. Hence, with the same pose pooling method, 2D consistency is more crucial than accuracy in establishing a stable training process. Ultimately, MagRep-based $\mathbf L^p$ induces the best PAMPJPE and 3D Acc. 

\begin{table}[t]
    \small
    \renewcommand{\arraystretch}{1.1}
    \setlength\tabcolsep{2pt}
    \centering
    \begin{tabular}{c  c c c c c}
    \hline
    3D decoding & Mult-Adds & \#Param & PJ$\downarrow$ & Acc$\downarrow$  & FPS$\uparrow$ \\
    \hline

    \multicolumn{6}{c}{ \scriptsize{\textit{w/ GhostStack}} } \\
    SpiralConv++  & 159.0/263.1M & 1.0/6.2M & 8.63 & 2.31/7.07 & 77 \\
    DSConv (ours) & 19.5/123.5M & 0.1/5.3M & 8.76 & 2.30/6.98 & \bf 83 \\

    \cdashline{1-6}[0.8pt/2pt]

    \multicolumn{6}{c}{ \scriptsize{\textit{w/ DenseStack}} } \\
    SpiralConv++   & 159.0/579.4M & 1.0/9.0M & \bf 6.85  & 1.98/4.75 & 59 \\
    DSConv (ours)  & 19.5/439.9M  & 0.1/8.1M & 6.87 & \bf 1.92/\bf 4.73 & 67  \\
    
    \hline
    \end{tabular}
    \caption{Ablation study of 3D decoding. Mult-Adds and \#Param are \textit{w.r.t.} the 3D decoder/overall model; 2D/3D Acc is presented; FPS is tested on Apple A14 CPU; Accuracy and temporal performance are tested on FreiHAND and HO3Dv2, respectively.}
    \label{tab:dsconv}
    \vspace{-0.3cm}
    \end{table}

\begin{figure*}[t]
\begin{center}
\includegraphics[width=\linewidth]{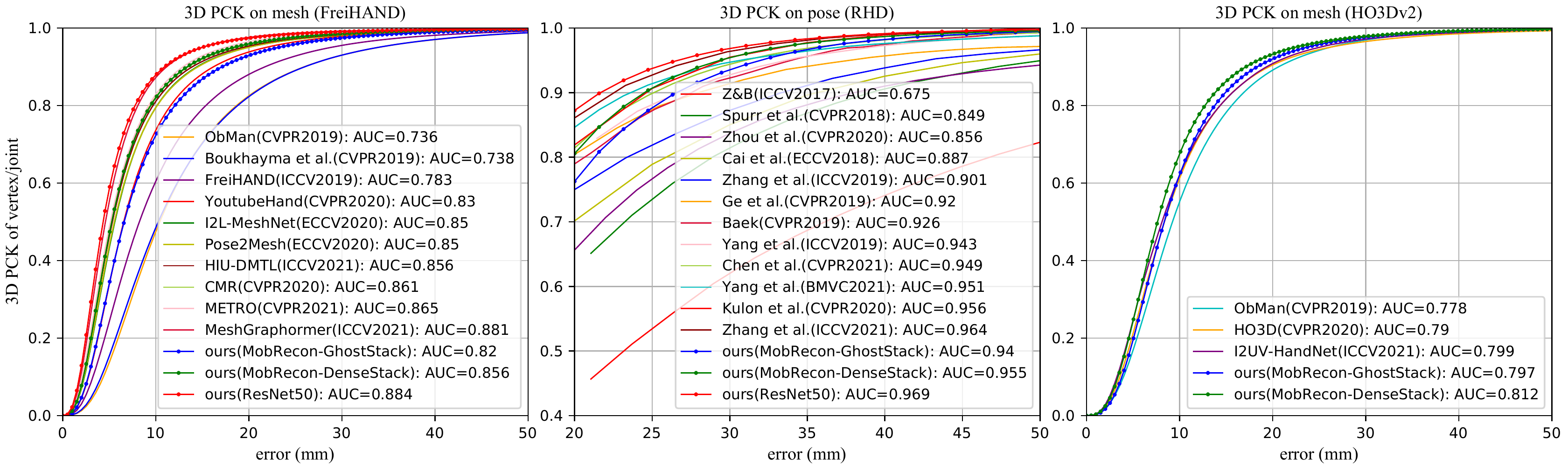}
\caption{3D PCK vs. error thresholds.}
\label{fig:curve}
\end{center}
\vspace{-0.5cm}
\end{figure*}

In PVL, we design a linear operation with lifting matrix $\mathbf M^l\in\mathbb R^{V^{mini}\times N}$ to transform features from 2D pose space to 3D vertex space. Thus, $V^{mini}$ vertex features are produced by a linear combination of $N$ landmark features. Figure~\ref{fig:matrix} depicts a well-trained lifting matrix, where we illustrate $\mathrm{abs}(\mathbf M^l)$ to clearly reveal the joint-vertex relations. As can be observed, the learned $\mathbf M^l$ is sparse. A joint landmark (\emph{i.e.}, joint 5 that locates at the root of forefinger) serves as the global information and contributes to the majority of vertices. Besides, some joint landmark traits are propagated to their corresponding vertices. Figure~\ref{fig:lifting} depicts highly relevant connections in $\mathbf M^l$, demonstrating that the PVL approach can preserve the semantic consistency. 

When compared to CMR \cite{bib:CMR} (the first row of Table~\ref{tab:2d}), our feature lifting module results in better PAMPJPE and 3D Acc. Moreover, we significantly reduce the computational expenses in the 2D-to-3D part. Despite the use of extra Mult-Adds in the 2D part, 2D pose prediction has previously proven to be beneficial in 3D hand reconstruction due to multi-task learning \cite{bib:HIU} and root recovery task \cite{bib:CMR}.

\vspace{-0.4cm}
\paragraph{Towards balancing model efficiency and performance.}

\begin{table}[t]
\small
\renewcommand{\arraystretch}{1.}
\setlength\tabcolsep{2.5pt}
\centering
\begin{tabular}{c c | c c c c}
\hline
Method  &  Backbone  & PJ $\downarrow$ & PV $\downarrow$ & F@5 $\uparrow$ & F@15 $\uparrow$ \\
\hline
MobileHand \cite{bib:MobileHand}   & MobileNet & $-$ & 13.1 & 0.439 & 0.902  \\
FreiHAND \cite{bib:FreiHAND}      & ResNet50 & 11.0 & 10.9 & 0.516 & 0.934  \\
YotubeHand \cite{bib:YoutubeHand} & ResNet50  & 8.4  & 8.6  & 0.614 & 0.966  \\
I2L-MeshNet \cite{bib:I2L}        & ResNet50$^*$ & 7.4 & 7.6   & 0.681 & 0.973  \\
HIU-DMTL \cite{bib:HIU}           & Customized$^*$ & 7.1 & 7.3   & 0.699 & 0.974  \\
CMR \cite{bib:CMR}                & ResNet50$^*$ & 6.9  & 7.0 & 0.715 & 0.977 \\   
I2UV-HandNet \cite{bib:I2UV}      & ResNet50           & 6.7  & 6.9 & 0.707 & 0.977 \\ 
METRO \cite{bib:Metro}  & HRNet & 6.7  & 6.8 & 0.717 & 0.981   
\\
Tang \etal \cite{bib:handAR}      & ResNet50           & 6.7  & 6.7 & 0.724 & 0.981 \\   
MeshGraphormer \cite{bib:MeshGraphormer}  & HRNet & 5.9  & 6.0 & 0.765 & \bf  0.987   
\\
\hline
MobRecon (ours)      & GhostStack$^*$& 8.8  & 9.1 & 0.597 & 0.960   \\
MobRecon (ours)      & DenseStack$^*$& 6.9  & 7.2 & 0.694 & 0.979   \\
ours$^\ddagger$     & ResNet18$^*$  & 6.7  & 6.8 & 0.727 & 0.979   \\
ours$^\dagger$            & ResNet18$^*$  & 6.1  & 6.3 & 0.758 & 0.983   \\
ours$^\ddagger$     & ResNet50$^*$  & 6.1  & 6.2 & 0.760 & 0.984   \\
ours$^\dagger$            & ResNet50$^*$  & \bf 5.7  & \bf 5.8 & \bf 0.784 & 0.986   \\
\hline
\end{tabular}
\caption{Results on the FreiHAND dataset. $^*$: stacked structure; $^\dagger$: These models are based on ImageNet pre-trained backbone and mixed fine-tuning data;$^\ddagger$: These models are totally unrelated to our complement data.}
\label{tab:freihand}
\vspace{-0.3cm}
\end{table}

We design 3D/2D consistency loss to improve the performance further. From Table~\ref{tab:consist}, we can see that consistency learning improves temporal coherence. Besides, the accuracy and temporal coherence can benefit from each other so that PAMPJPE is also enhanced.

As shown in Table~\ref{tab:dsconv}, DSConv dramatically decreases the Mult-Adds and \#Param of the 3D decoder and obtains on par, sometimes even better, performance compared with SpiralConv++. Overall, our MobRecon with DenseStack/GhostStack can reach 67/83 FPS on Apple A14 CPU.

\vspace{-0.4cm}
\paragraph{Discussion.}
MobRecon has a limitation that the DSConv increases memory access cost, so some engineering optimization should be involved for higher inference speed.

\subsection{Comparisons with Contemporary Methods}

On the FreiHAND dataset, we scale up our ResNet-based model with $224\times 224$ input resolution for a fair comparison. As shown in Table~\ref{tab:freihand}, we surpass previous methods with ResNet50, leading to a new state of the art, \emph{i.e.}, 5.7mm PAMPJPE. Based on DenseStack or GhostStack, our MobRecon outmatches some ResNet-based methods. Referring to Figure~\ref{fig:curve}, the proposed MobRecon has superior performance on 3D PCK. Beyond high accuracy, we also achieve superior inference speed, as shown in Figure~\ref{fig:tradeoff}.

In experiments on RHD and HO3Dv2, our complement data are only used to pre-train DenseStack/GhostStack. On the RHD dataset, we compare with several pose estimation methods (such as \cite{bib:Baek19,bib:Cai18,bib:Spurr18,bib:Yang19})
in Figure~\ref{fig:curve}.
Our MobRecon with DenseStack/GhostStack induces 3D AUC of 0.955 and 0.940, outperforming most compared approaches.

\begin{table}[t]
\small
\renewcommand{\arraystretch}{1.}
\setlength\tabcolsep{2pt}
\centering
\begin{tabular}{c c | c c c c}
\hline
Method  &  Backbone   & PJ $\downarrow$ & PV $\downarrow$ & F@5 $\uparrow$ & F@15 $\uparrow$ \\
\hline
ObMan \cite{bib:ObMan} & ResNet18  & 11.0 & 11.0 & 0.464 & 0.939  \\
HO3D \cite{bib:HO3D}   & CPM \cite{bib:CMP}  & 10.7 & 10.6 & 0.506 & 0.942  \\
I2UV-HandNet \cite{bib:I2UV}   & ResNet50  & 9.9 & 10.1 & 0.500 & 0.943  \\
\hline
MobRecon (ours) & GhostStack & 10.0 & 10.2 & 0.488 & 0.948   \\
MobRecon (ours) & DenseStack & \bf 9.2  & \bf 9.4  & \bf 0.538  & \bf 0.957   \\
\hline
\end{tabular}
\caption{Results on the HO3Dv2 dataset.}
\label{tab:ho3d}
\vspace{-0.3cm}
\end{table}

The HO3Dv2 dataset is employed for evaluation. According to Table~\ref{tab:ho3d}, our MobRecon outperforms existing methods such as \cite{bib:ObMan,bib:HO3D,bib:I2UV}. HO3Dv2 is more challenging than FreiHAND and RHD because of serious object occlusion. In this case, MobRecon outperforms some ResNet-based methods because of better generalization ability. Besides, we also achieve better temporal coherence in this sequential task, as shown in Table~\ref{tab:2d}.

Please refer to the supplementary materials for more qualitative analyses, mobile applications, \textit{etc.}

\section{Conclusions and Future Work}

In this work, we present a novel hand mesh reconstruction method with superior efficiency, accuracy, and temporal coherence. First, we propose lightweight stacked structures for 2D encoding. Then, a feature lifting module with MapReg, pose pooling, and PVL approaches is designed for 2D-to-3D mapping. Besides, DSConv is developed to handle the 3D decoding task efficiently. Our MobRecon only involves 123M Mult-Adds and 5M parameters so that it reaches a fast inference speed of 83FPS on Apple A14 CPU. Moreover, we achieve the state-of-the-art performance on FreiHAND, RHD, and HO3Dv2. We plan to investigate efficient methods for interacting hands.

{\small

}

\newpage

\section{Complement Dataset}
\label{sec:data}

\paragraph{Motivation.}
As shown in Table~\ref{tab:dataset}, many datasets are developed for 3D hand pose estimation \cite{bib:STB,bib:RHD,bib:EgoDexter,bib:DexterObject,bib:FreiHAND,bib:YoutubeHand,bib:MHP,bib:H3D,bib:DexYCB,bib:H2O,bib:FPHA,bib:SeqHAND,bib:GANerated}. To collect real-world hand data, existing datasets are usually captured using a multi-view studio and annotated via semi-automatic model fitting \cite{bib:FreiHAND,bib:MHP}. However, these model-fitted datasets usually suffer from noisy annotation, lack of background diversity, and costly data collection. An alternative way is computer-aided synthetic data \cite{bib:RHD,bib:MVHM}, which are superior in scalability, distribution, annotation, and collection cost. In addition, a good training dataset should avoid long-tailed distributions. That is, both hand poses and viewpoints should be uniformly distributed. Unfortunately, we are not aware of any existing dataset that fits this requirement. Some datasets try to alleviate the problem of limited viewpoints by multi-view rendering (\textit{e.g.}, MVHM \cite{bib:MVHM} contains 8 views), but they are still too sparse to cover all the possible views. Boukhayma \etal \cite{bib:Boukhayma19} uniformly sampled MANO PCA components to produce various hand poses. However, the PCA space does not describe physical factors, so the corresponding sampling results cannot be intuitively controlled. Thereby, we are inspired to generate a more comprehensive hand dataset with sufficient and uniformly distributed hand poses and viewpoints.

\begin{table}
\small
\renewcommand{\arraystretch}{1.}
\setlength\tabcolsep{3.2pt}
\centering
\begin{tabular}{c | c c c c c c}
\hline
Dataset                  & Type & Size  & Mesh & UP & MV \\
\hline
STB \cite{bib:STB}       & real    & 36K  & $\times$ & $\times$  & $\times$   \\
RHD \cite{bib:RHD}       & synthetic & 44K  & $\times$ & $\times$ & $\times$   \\
GANerated Hands \cite{bib:GANerated} & synthetic & 331K  & $\times$ & $\times$ & $\times$ \\
SeqHAND \cite{bib:SeqHAND} & synthetic & 410K  & $\checkmark$ & $\times$ & $\times$ \\

EgoDexter \cite{bib:EgoDexter} & real & 3K  & $\times$ & $\times$ & $\times$  \\
Dexter+Object \cite{bib:DexterObject} & real & 3K & $\times$ & $\times$ & $\times$   \\
FreiHAND \cite{bib:FreiHAND}  & real & 134K & $\checkmark$ & $\times$ & $\times$   \\
YoutubeHand \cite{bib:YoutubeHand} & real & 47K  & $\checkmark$ & $\times$ & $\times$   \\
ObMan \cite{bib:ObMan} & synthetic & 153K & $\checkmark$ & $\times$ & $\times$   \\
HO3D \cite{bib:HO3D} & real & 77K & $\checkmark$ & $\times$ & $\times$    \\
DexYCB \cite{bib:DexYCB} & real & 528K & $\checkmark$ & $\times$ & $\times$    \\
H2O \cite{bib:H2O} & synthetic & 571K & $\checkmark$ & $\times$ & $\times$   \\
FPHA \cite{bib:FPHA} & synthetic & 105K & $\times$ & $\times$ & $\times$   \\
H3D \cite{bib:H3D}   & real & 22K & $\checkmark$ & $\times$ & $\checkmark$(15) \\
MHP \cite{bib:MHP} & real & 80K  & $\times$  & $\times$ & $\checkmark$(4)  \\
MVHM \cite{bib:MVHM} & synthetic  & 320K  & $\checkmark$ & $\times$  & $\checkmark$(8) \\
InterHand2.6M \cite{bib:InterHand} & real & 2.6M & $\checkmark$ & $\times$ & $\checkmark$(80)  \\
\hline
ours & synthetic  & 328K & $\checkmark$ & $\checkmark$ & $\checkmark$(216)   \\
\hline
\end{tabular}
\caption{Comparison among RGB-based 3D hand datasets. ``MV'' means multi-view, and the number in brackets shows the total number of views. ``UP'' denotes uniform pose distribution.}
\label{tab:dataset}
\vspace{-0.2cm}
\end{table}

\begin{figure}[t]
    \begin{center}
    \includegraphics[width=\linewidth]{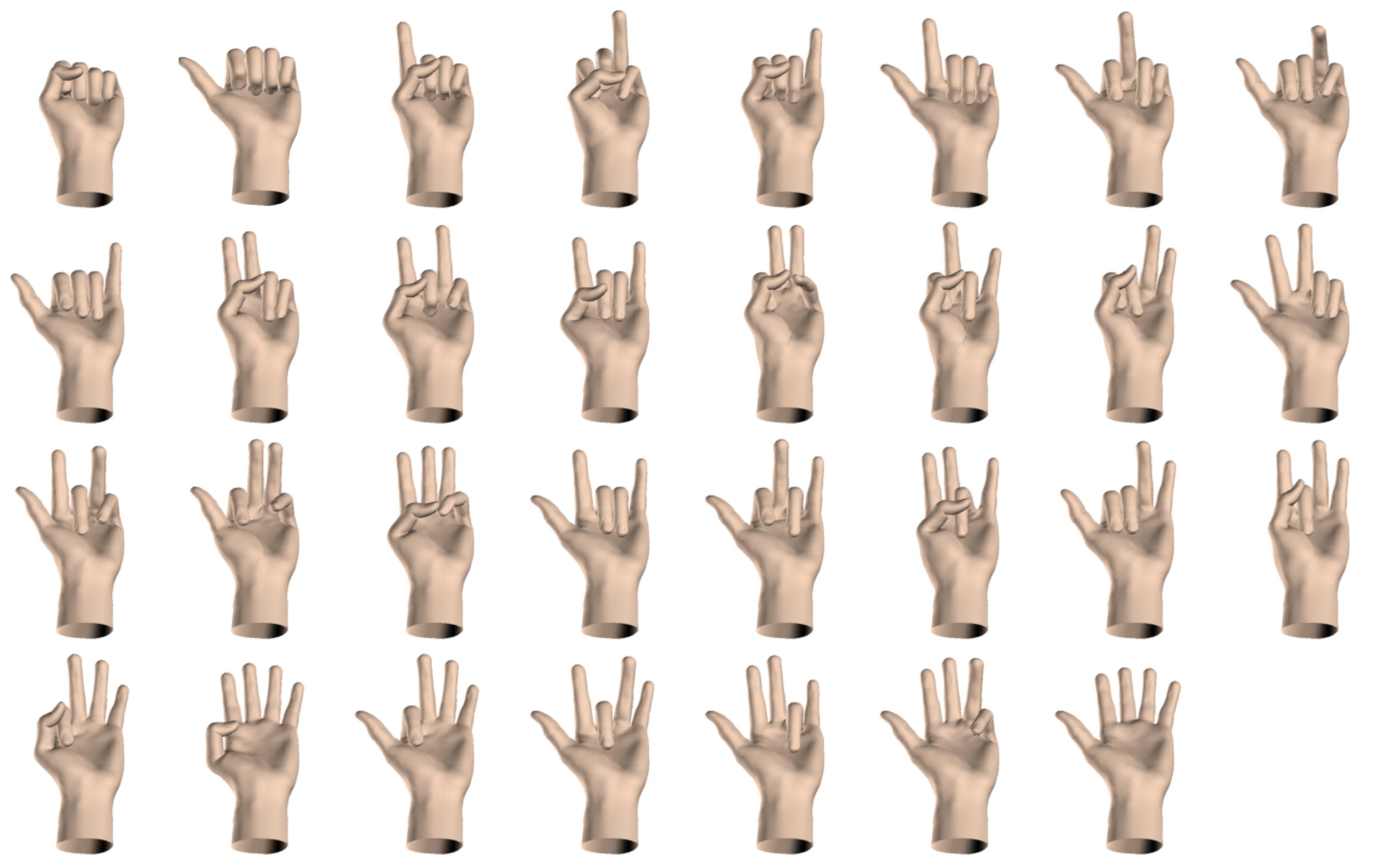}
    \caption{Base poses. Under the consideration of politeness, one pose with middle finger extending is not shown.}
    \label{fig:base_pose}
    \end{center}
    \vspace{-0.5cm}
    \end{figure}

\begin{figure}[t]
\begin{center}
\includegraphics[width=0.95\linewidth]{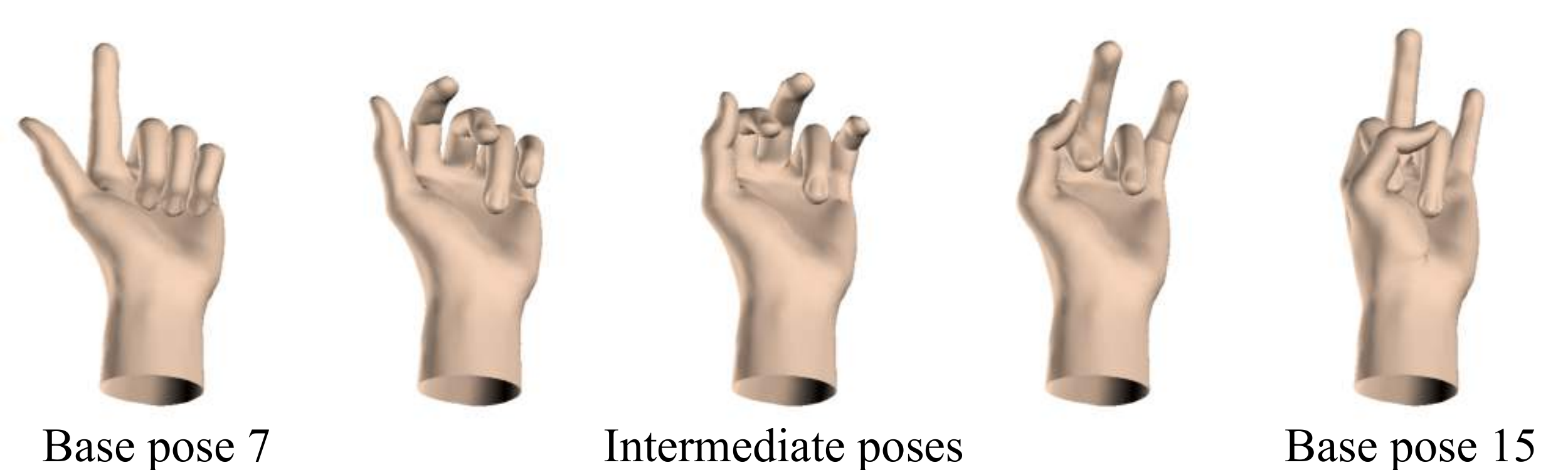}
\caption{Intermediate poses from base pose 7 to base pose 15.}
\label{fig:trans_pose}
\end{center}
\vspace{-0.3cm}
\end{figure}

\vspace{-0.3cm}
\paragraph{Data Designs.}
We design a high-fidelity hand mesh with $5633$ vertices and $11232$ faces. Different from existing hand datasets, we uniformly design hand poses. First, as shown in Figure~\ref{fig:base_pose}, we set two states for each finger, \emph{i.e.}, total bending and extending. Then, we obtain 32 base poses by combining five finger states. The combination of these base poses results in 496 pose pairs. For each pair, we uniformly interpolate three intermediate poses from one pose to another in Maya software\footnote{https://www.autodesk.com/} (as shown in Figure~\ref{fig:trans_pose}). In total, we obtain 1520 uniformly distributed hand poses.

\begin{figure}[t]
\begin{center}
\includegraphics[width=0.7\linewidth]{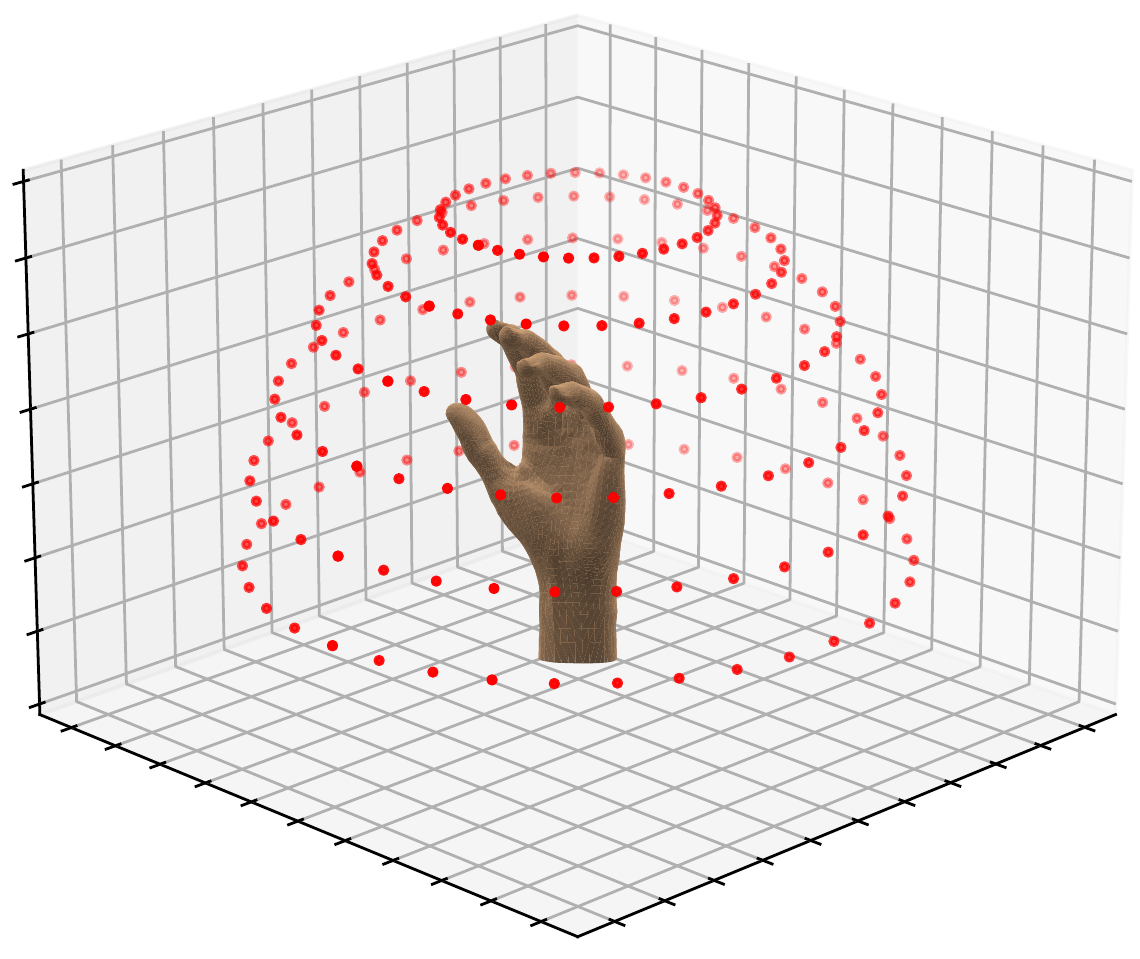}
\caption{Illustration of viewpoints to render the dataset. Each red point denotes a camera position.The camera points to the palm center.}
\label{fig:pano_point}
\end{center}
\vspace{-0.5cm}
\end{figure}

\begin{figure}[t]
\begin{center}
\includegraphics[width=0.95\linewidth]{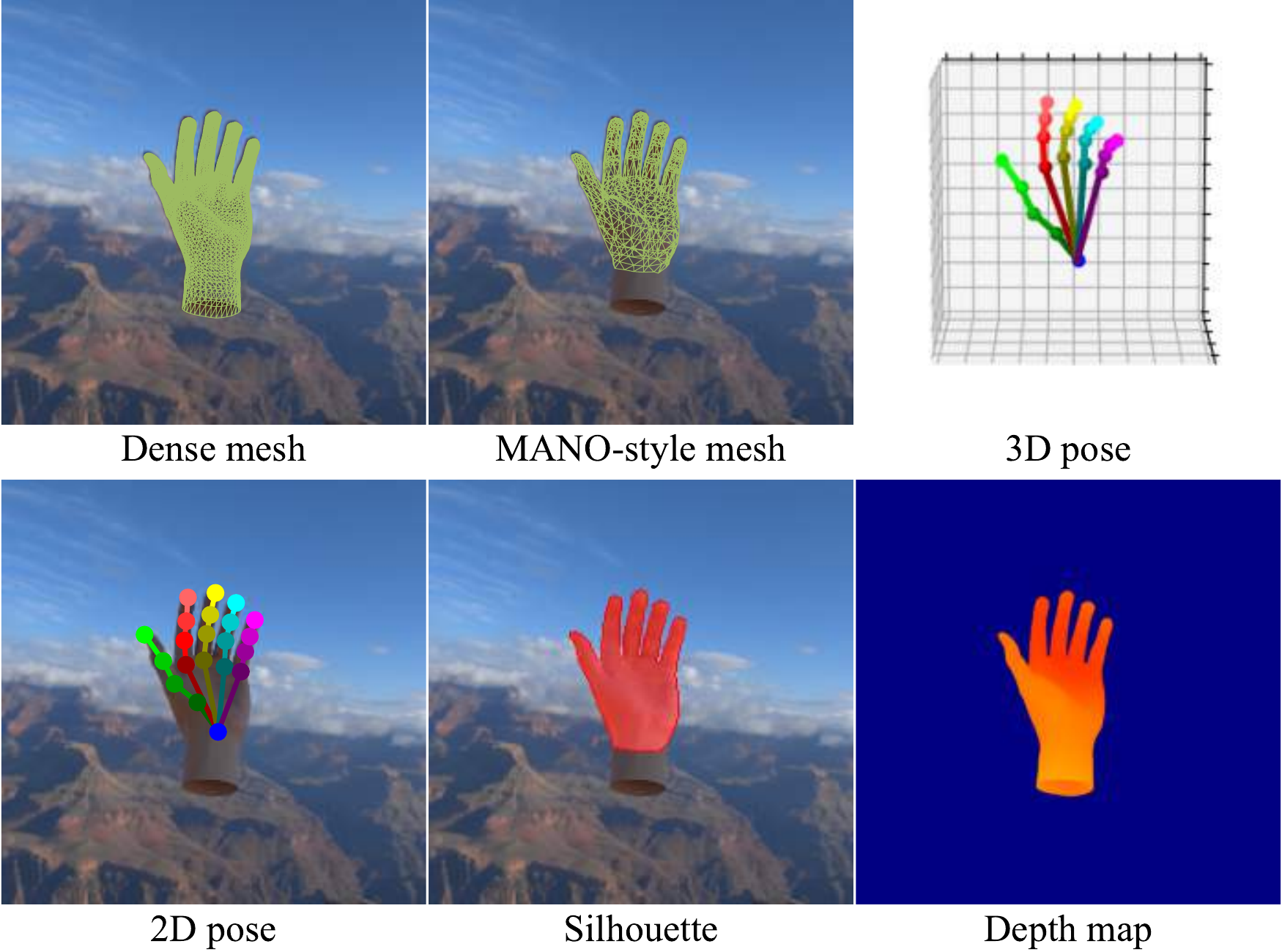}
\caption{Annotations.}
\label{fig:anno}
\end{center}
\vspace{-0.5cm}
\end{figure}

For each pose sample, we provide its dense viewpoints by rendering. To this end, we uniformly define 216 hemispherical-arranged camera positions. As shown in Figure~\ref{fig:pano_point}, the longitude ranges from $0$ to $2\pi$ while the latitude ranges from $0$ to $\pi/2$. Adjacent positions differ in longitude or latitude by $\pi/18$ or $\pi/12$. All cameras point to the palm center so that the hand locates at the center of rendered images. Because the end of the wrist locates at the sphere center, the hemispherical sampling contains the first-person perspectives. As for the background, we collect high-dynamic-range (HDR) imaging with real scenes and illumination for rendering so that our hand mesh can realistically blend into various scenes. Figure~\ref{fig:pano} illustrates rendered samples with our viewpoints.

The automatically generated annotations involve no noise. Consistent with mainstream datasets, we design a pose-agnostic matrix to map our dense topology to MANO-style mesh with 778 vertices and 1538 faces. As shown in Figure~\ref{fig:anno}, we provide annotations of our designed dense mesh, MANO-style mesh, 3D pose, 2D pose, silhouette, depth map, and intrinsic camera parameters.

\vspace{-0.3cm}
\paragraph{Discussion.}
The limitation of our data is the lack of shape/texture diversity. Additionally, we only consider finger bend, and we plan to model finger splay to extend this dataset to cover the entire pose space uniformly.

\vspace{-0.3cm}
\paragraph{Network pre-training.}

\begin{figure}[t]
\begin{center}
\includegraphics[width=\linewidth]{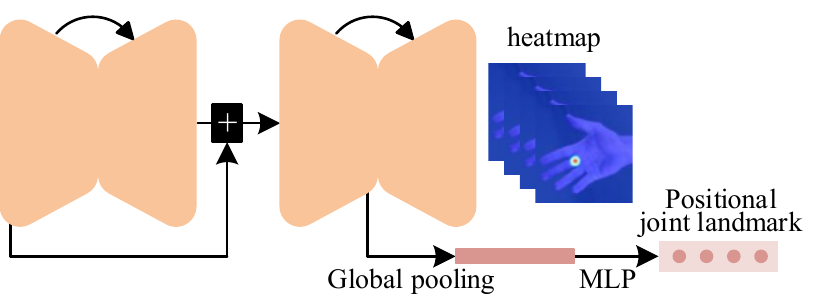}
\caption{The pre-training architecture using our data.}
\label{fig:pretrain}
\end{center}
\vspace{-0.cm}
\end{figure}
To pre-train the 2D encoding network, we design a 2D pose estimation task without the need of 3D annotation. In the main text, we analyzes 2D representations with heatmap and position regression. Hence, as shown in Figure~\ref{fig:pretrain}, we equally consider these representations during the pre-training step. That is, both heatmap and positional joint landmark are supervised. The model is pre-trained for 80 epochs with a mini-batch size of 128. The initial learning rate is $10^{-3}$, which is divided by 10 at the 20th, 40th, and 60th epochs. The input resolution is $128\times 128$.

\section{Analysis and Application}

\begin{figure}[t]
\begin{center}
\includegraphics[width=0.8\linewidth]{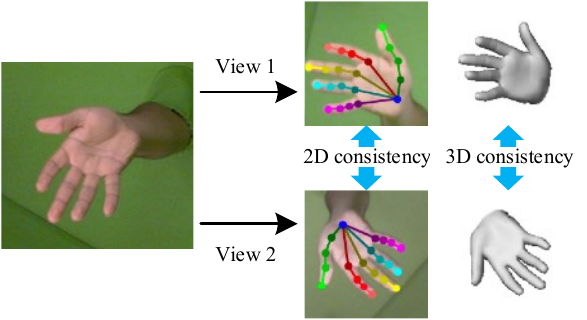}
\caption{Consistency loss based on data augmentation.}
\label{fig:consist}
\end{center}
\vspace{-0.55cm}
\end{figure}
\paragraph{Diagram of our consistency loss.}
As shown in Figure~\ref{fig:consist}, two views are derived with data augmentation with an input image. Then consistency loss can be designed in both 2D and 3D spaces as Equation 11 in the main text.

\vspace{-0.3cm}
\paragraph{Explanation of dataset setting.}
During the ablation study in the main text, we use RHD, FreiHAND, and HO3Dv2 to evaluate different properties.
Because FreiHAND and HO3Dv2 do not release ground truth and the official tools do not support 2D evaluation, RHD is employed for testing 2D accuracy. 
HO3Dv2 is a sequential dataset, so it is adopted to reflect temporal coherence.
However, HO3Dv2 highlights hand-object interaction, which is not our topic. In contrast, FreiHAND highlights various hand poses, lighting conditions, \textit{etc.}, so we use it for evaluating 3D accuracy.

\vspace{-0.3cm}
\paragraph{The effect of our complement data during fine-tuning.}

\begin{figure}[t]
\begin{center}
\includegraphics[width=\linewidth]{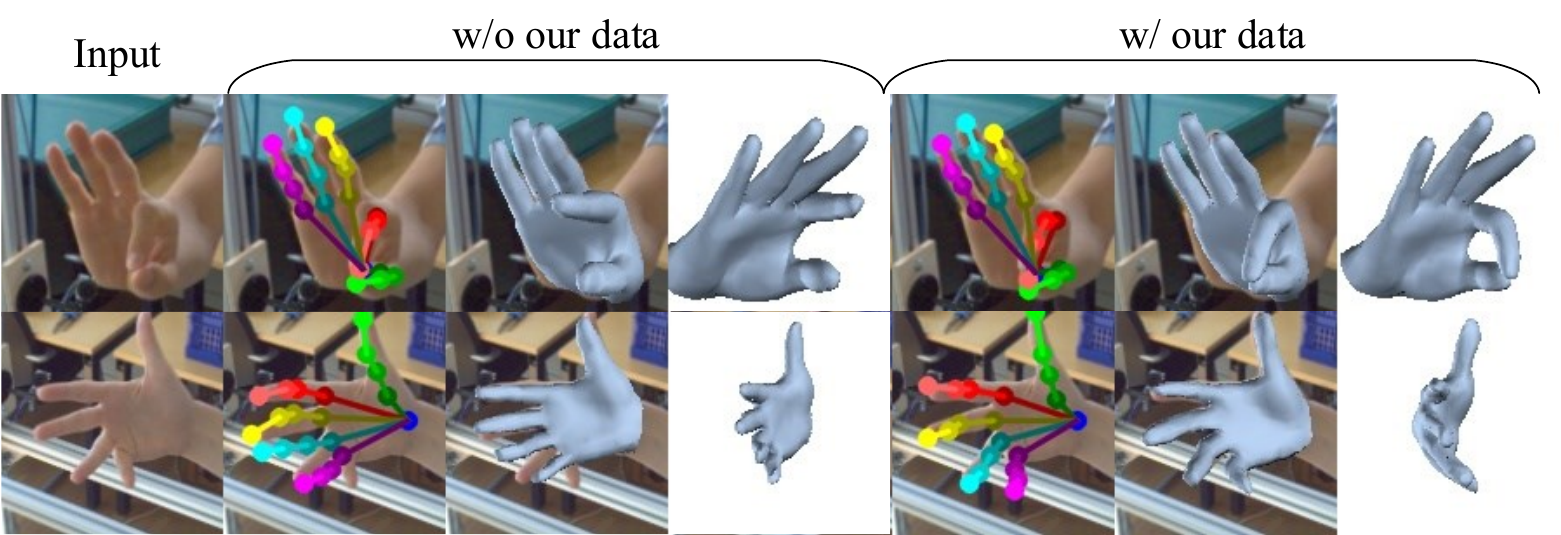}
\caption{Qualitative visualization of 2D pose, aligned mesh, and side-view mesh.}
\label{fig:vis}
\end{center}
\vspace{-0.3cm}
\end{figure}

As shown in Figure~\ref{fig:vis}, when our data are employed in fine-tuning step, it can improve the model performance on difficult pose prediction.

\vspace{-0.3cm}
\paragraph{Visualization on HO3Dv2.}

\begin{figure}[t]
\begin{center}
\includegraphics[width=\linewidth]{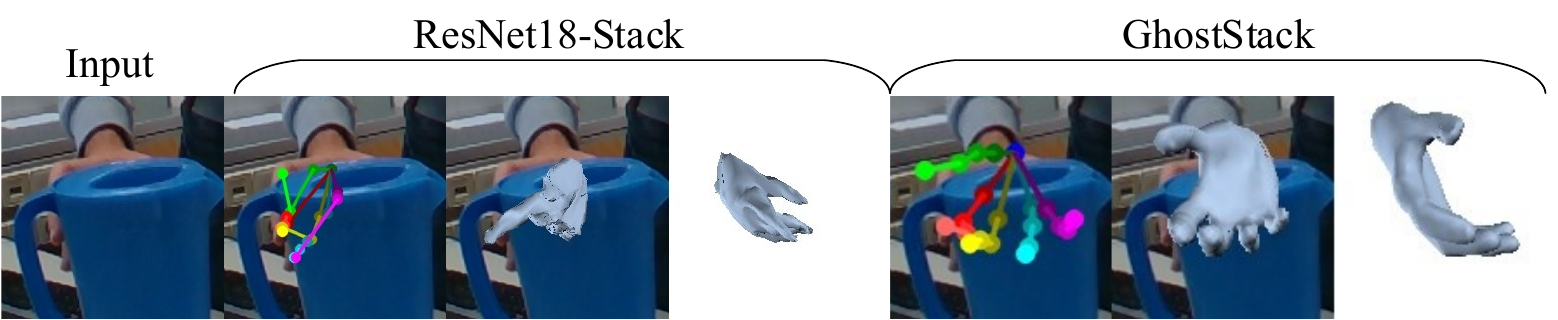}
\caption{Comparison of GhostStack and ResNet18 on a challenging HO3Dv2 sample.}
\label{fig:ho3d}
\end{center}
\vspace{-0.3cm}
\end{figure}

Referring to Table 6 in the main text, our MobRecon outperforms some ResNet-based models. We observe that this phenomenon is related to generalization performance. As shown in Figure~\ref{fig:ho3d}, HO3Dv2 contains massive seriously occluded samples. Under this extreme condition, our model can produce a physically correct prediction while the ResNet-based model collapses.

\begin{figure}[t]
\begin{center}
\includegraphics[width=\linewidth]{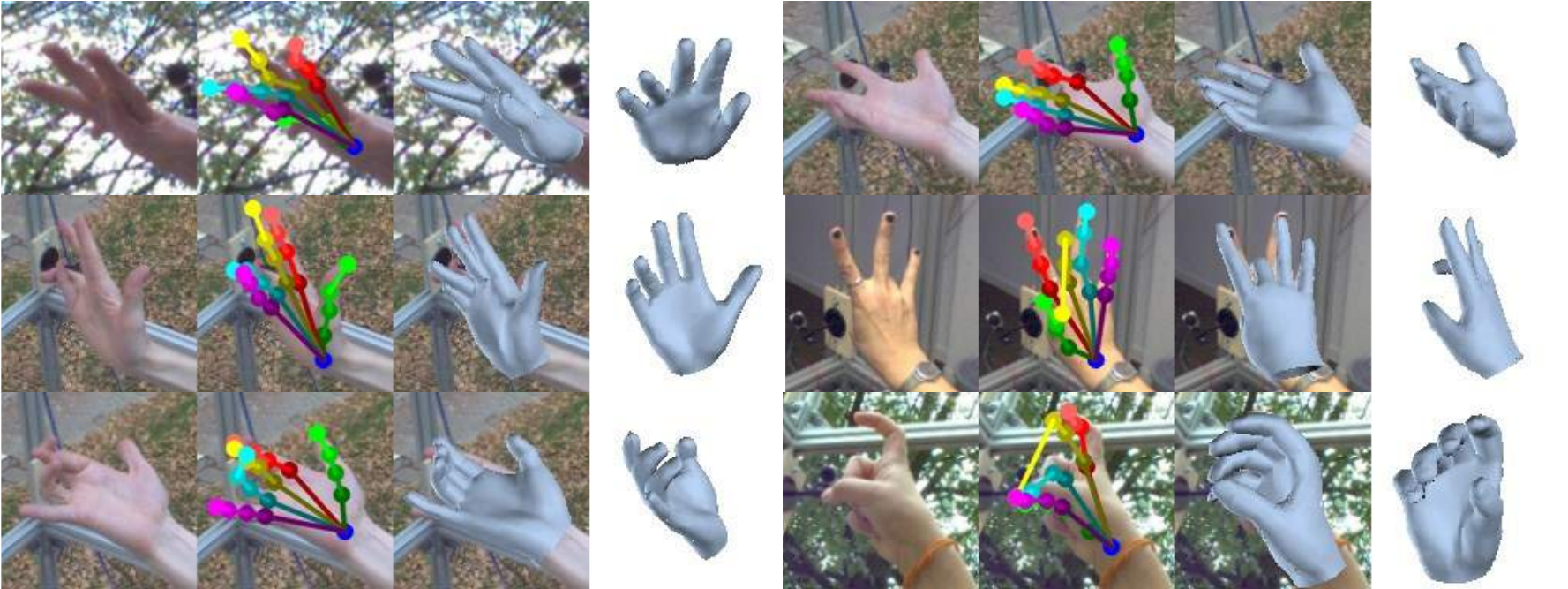}
\caption{Typical failure cases.}
\label{fig:bad}
\end{center}
\vspace{-0.3cm}
\end{figure}

\begin{figure}[t]
\begin{center}
\includegraphics[width=\linewidth]{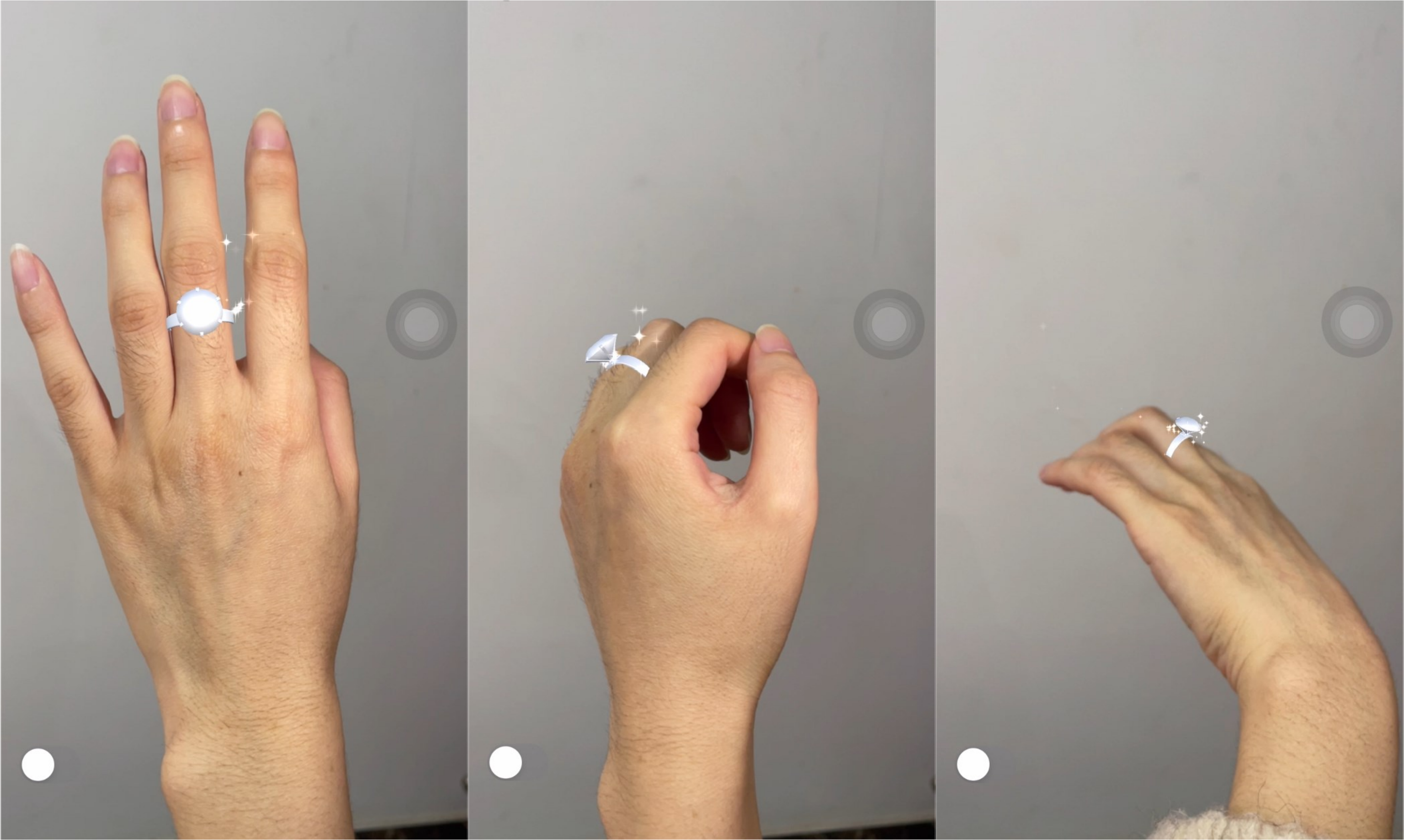}
\caption{We develop an AR effect with MobRecon and deploy it on mobile devices. This figure is captured with iPhone12.}
\label{fig:mobapp}
\end{center}
\vspace{-0.3cm}
\end{figure}

\vspace{-0.3cm}
\paragraph{Failure case analysis.}

As shown in Figure~\ref{fig:bad}, MobRecon could suffer from failure cases as for challenging poses. Typically, self-occlusion by finger splay is hard to accurately predict because they are tail-distributed poses in most datasets. We will solve this problem by improving our complement data, as stated in the above section.

\vspace{-0.3cm}
\paragraph{More qualitative results.}
Figure~\ref{fig:vis_supp} illustrates comprehensive qualitative results of our predicted 2D pose, aligned and side-view mesh. The challenges include challenging poses, object occlusion, truncation, and bad illumination. Overcoming these difficulties, our method can generate accurate 2D pose and 3D mesh.

\vspace{-0.3cm}
\paragraph{Qualitative comparison on temporal coherence.}
We record a video snippet to demonstrate temporal coherence, where we keep the camera and hand static to produce low acceleration. Despite the static condition, the network input could be temporally unstable because of detection jitter \etc. The ground-truth pose is straightforward (see Figure~\ref{fig:acc}), and all compared models can easily obtain high accuracy. Hence, temporal performance can be exclusively revealed in this experiment. As shown in Figure~\ref{fig:acc}, our MobRecon performs better than CMR \cite{bib:CMR} in terms of 2D/3D pose consistency. In addition, we also compute the root coordinates with the method in \cite{bib:CMR} and achieve better root recovery stability. Besides, we also complement 2D PCK curves on RHD, which demonstrate that our method has better 2D pose accuracy. Beyond accuracy and temporal coherence, our MobRecon with MapReg can produce better articulated structures because of global receptive field and adaptive inter-landmark constraints (see Figure 6 in the main text).

\vspace{-0.3cm}
\paragraph{Mobile application.}
Based on our MobRecon, a virtual ring can be worn with AR technique (Figure~\ref{fig:mobapp}).


\begin{figure*}[t]
\begin{center}
\includegraphics[width=0.95\linewidth]{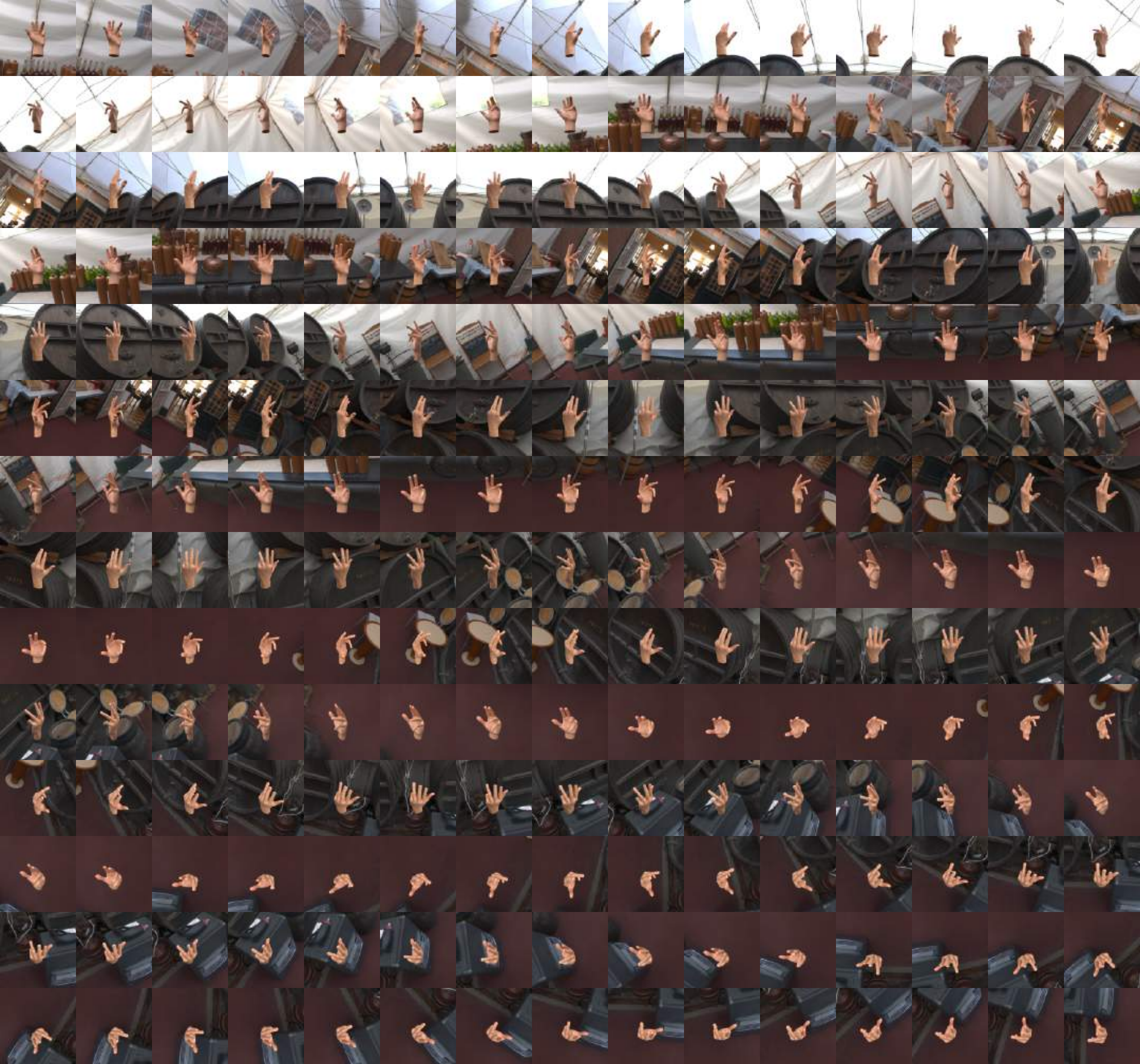}
\caption{Rendered samples with dense and uniform viewpoints.}
\label{fig:pano}
\end{center}
\vspace{-0.cm}
\end{figure*}

\begin{figure*}[t]
\begin{center}
\includegraphics[width=\linewidth]{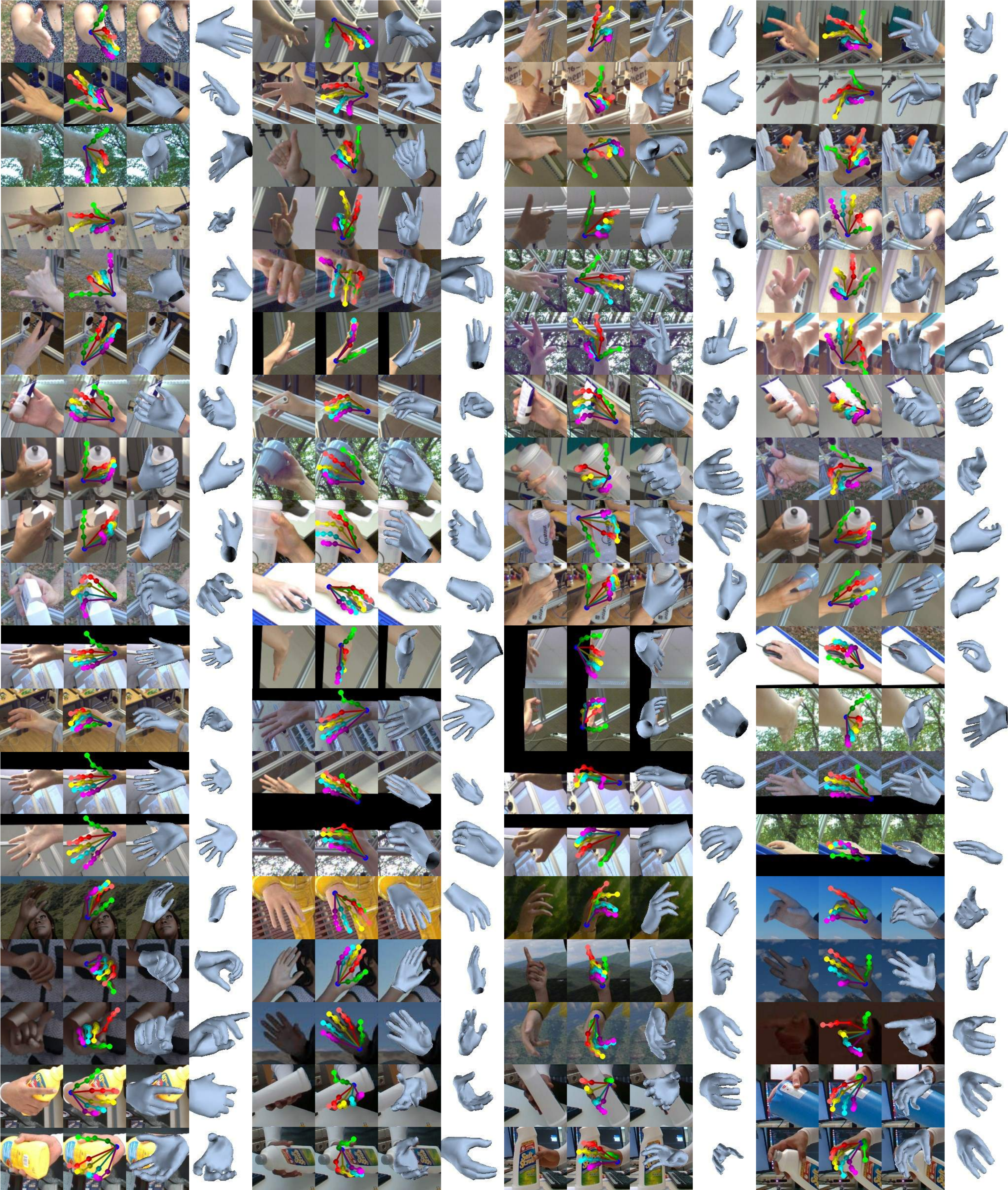}
\caption{Qualitative visualization of 2D pose, aligned mesh, and side-view mesh on FreiHAND, RHD, and HO3Dv2. Our method is robust enough to handle cases of occlusion, truncation, challenging poses, and bad illumination.}
\label{fig:vis_supp}
\end{center}
\end{figure*}

\begin{figure*}[t]
\begin{center}
\includegraphics[width=\linewidth]{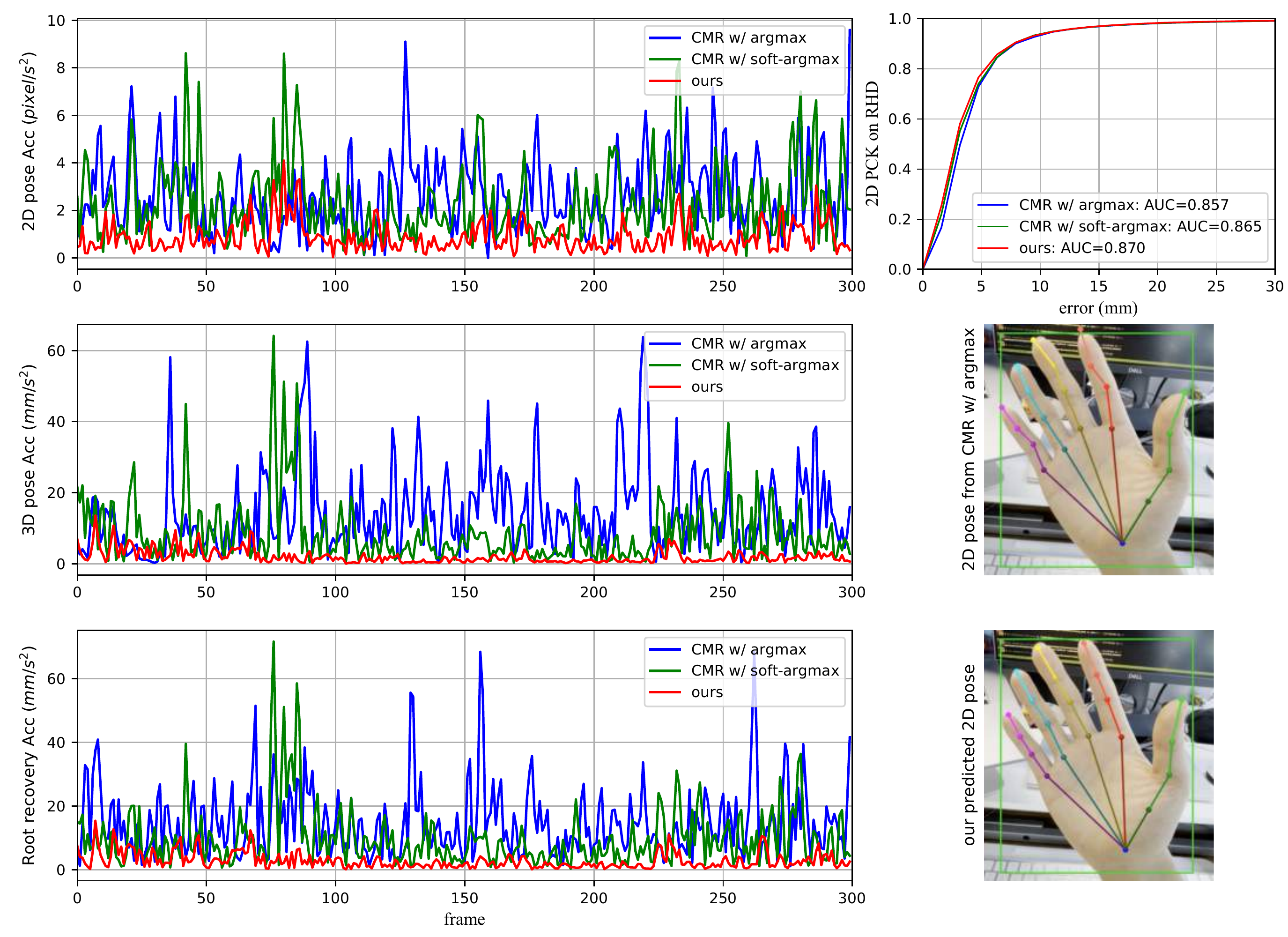}
\caption{We record a video snippet with a straightforward and static hand pose (see the bottom right corner) to compare the temporal performance and articulated structure.}
\label{fig:acc}
\end{center}
\vspace{-0.3cm}
\end{figure*}


\end{document}